\documentclass[11pt]{article}

\usepackage[preprint]{acl}

\usepackage{times}
\usepackage{latexsym}

\usepackage[T1]{fontenc}

\usepackage[utf8]{inputenc}

\usepackage{microtype}

\usepackage{inconsolata}

\usepackage{graphicx}
\usepackage{minted}
\usepackage{booktabs}
\usepackage{placeins}
\usepackage{listings}
\usepackage{bm}
\usepackage{enumitem}
\usepackage{tcolorbox}
\usepackage{multirow}

\title{RECOM: A Validity--Discrimination Tradeoff in Automatic Metrics for Open-Ended Reddit Question Answering}

\author{
 \textbf{Pushwitha Krishnappa\textsuperscript{1}},
 \textbf{Amit Das\textsuperscript{2}},
 \textbf{Vinija Jain\textsuperscript{3, 4}}\thanks{Work does not relate to position at Meta.},
 \\
 \textbf{Aman Chadha\textsuperscript{3, 5}}\thanks{Work does not relate to position at Amazon.},
 \textbf{Tathagata Mukherjee\textsuperscript{1}}
\\
\\
 \textsuperscript{1}University of Alabama Huntsville,
  \textsuperscript{2}University of North Alabama,\\
 \textsuperscript{3}Stanford University,
 \textsuperscript{4}Meta AI,
 \textsuperscript{5}Amazon GenAI
\\
\small{
  \textbf{Corresponding author:} Pushwitha Krishnappa (\href{mailto:email@domain}{pk0055@uah.edu})
 }
}

\begin{document}
\maketitle

\begin{abstract}
Automatic metrics are the default for evaluating LLM-generated text, yet a metric is quietly asked to do two jobs: tell genuine content alignment from surface coincidence (\emph{validity}), and tell a better system from a worse one (\emph{discriminative power}). On open-ended, opinion-driven question answering, the two are in tension. We introduce RECOM (Reddit Evaluation for Correspondence of Models), a contamination-free evaluation dataset of 15{,}000 r/AskReddit questions (September 2025), each paired with its authentic community replies, which postdate every evaluated model's training cutoff. Scoring five open-source LLMs (7--10B) against every reply each metric paired with a random-derangement noise floor we find that no metric does both jobs well. Cosine similarity separates real from random answers (Cohen's $d \approx 2$) but cannot rank the five models ($|d| < 0.1$); BERTScore precision \emph{appears} to rank the models (raw $|d|$ up to 0.63), but once response length is controlled this collapses to $|d| = 0.09$ and its validity is weak ($d \approx 0.8$, versus cosine's $\approx 2$). Because every metric scores the same outputs, this validity--discrimination tradeoff is a property of the metrics, not the models, and we argue it stems from representation design. Three independent LLM judges reproduce the validity gap and likewise separate the five models only weakly. We recommend reporting metrics on both axes, with an explicit random-baseline floor. RECOM is publicly available at \url{https://anonymous.4open.science/r/recom-D4B0}
\end{abstract}

\section{Introduction}

Large Language Models (LLMs) are increasingly used to answer open-ended, opinion-driven questions in conversational agents, search assistants, and question-answering systems where no single answer is correct. Evaluating such answers relies on automatic metrics, yet how standard metrics behave on open-ended, opinion-driven tasks remains poorly understood.

A metric used for evaluation is implicitly asked to do two distinct jobs. The first is a matter of \emph{validity}: does a high score reflect genuine alignment with the reference, or could it equally arise from surface coincidence between unrelated fluent text? The second is a matter of \emph{discriminative power}: given two systems, does the metric reliably separate the better one from the worse? These are routinely treated as a single property a ``good metric'' yet they need not coincide. Consider a researcher who evaluates an LLM on open-ended questions, observes a BERTScore F1 of 86\%, and concludes that the model is strongly aligned with human references. That score may sit only two points above the value expected for random English text pairs under the same embedding model a validity gap that is invisible without baseline rescaling. Now suppose the researcher instead adopts a metric that \emph{does} cleanly separate real answers from random ones, such as embedding cosine similarity. We show that on open-ended questions this metric assigns five different LLMs almost identical scores: it has validity but almost no discriminative power. With a single metric, the researcher cannot both trust that the score is meaningful and use it to choose between models.

This is not an artifact of two hand-picked metrics. Across a portfolio of lexical, semantic, and inference-based metrics, we find a systematic \emph{validity--discrimination tradeoff}: the metrics that best separate genuine alignment from noise are the weakest at ranking systems, and the metric that separates systems best opens only a narrow margin over noise and even that separation, we find, is largely an artifact of response length. No metric we evaluate does both well.

Demonstrating this tradeoff calls for an evaluation setting that is open-ended by construction and free of training-data contamination. Recent work has begun to address temporal robustness in LLM evaluation: FreshQA \citep{vu2023freshllms}, RealTimeQA \citep{kasai2024realtime}, and TimeQA \citep{chen2021timeqa} test LLMs on factoid questions with time-sensitive, objectively verifiable answers. However, real-world user questions, especially those posed on online platforms such as Reddit, are often open-ended and opinion-driven, where ``correctness'' is better understood as alignment with community consensus rather than factual accuracy.

\begin{figure*}
\centering
  \includegraphics[width=1\linewidth]{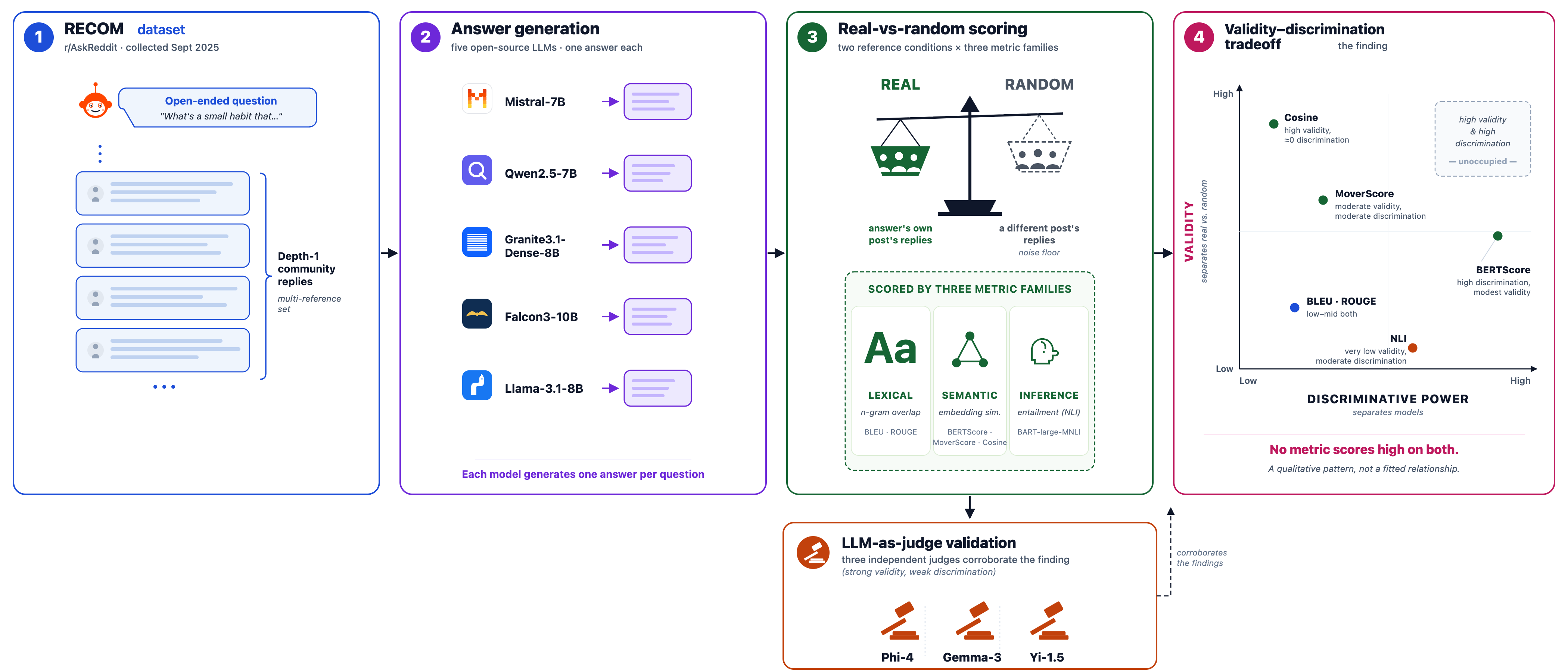} 
  \caption{Overview of the RECOM evaluation pipeline: (i) extract Reddit questions and their depth-1 community replies; (ii) generate answers with multiple LLMs; (iii) evaluate each generated answer against the community replies using lexical, semantic, inference-based, and LLM-as-judge metrics.}
  \label{fig:workflow}
\end{figure*}

In this work, we turn open-ended Reddit question answering into a testbed for metric behavior. For each question we take its depth-1 community replies the genuine human answers as a multi-reference set, and score the answers of five open-source LLMs against every reply using lexical, semantic, and inference-based metrics (Figure~\ref{fig:workflow}). Pairing every metric with a random-derangement baseline lets us measure two quantities on a common scale (Cohen's $d$): a \emph{validity} score, the separation between real and random generations, and a \emph{discriminative-power} score, the separation between models. Plotting metrics in this two-dimensional space exposes the tradeoff (Figure~\ref{fig:tradeoff}), and an LLM-as-judge validation with three independent judges corroborates the validity axis.

Our contributions are as follows:
\begin{enumerate}[nosep,leftmargin=*]
\item We identify and quantify a \textbf{validity--discrimination tradeoff} in automatic metrics for open-ended QA: among the metrics we evaluate, none both separates genuine alignment from noise and ranks systems reliably. Cosine similarity maximizes validity (real-vs-random Cohen's $d \approx 2$) but has almost no discriminative power ($|d| < 0.1$); BERTScore precision does the reverse, though residualizing response length shows that discriminative power is largely a length effect ($|d|$ from 0.63 down to 0.09). Because every metric scores the \emph{same} model outputs, the tradeoff is a property of the metrics, not the models.
\item We introduce \textbf{RECOM}, an evaluation dataset of 15{,}000 open-ended English-language r/AskReddit questions, each paired with its authentic depth-1 community replies and drawn from a contamination-free temporal window a testbed with a built-in random-baseline floor that makes both metric axes measurable (Figure~\ref{fig:workflow}).
\item We put forward a \textbf{representation-design hypothesis}: that contrastively-trained sentence encoders sharpen the real-vs-random contrast at the cost of the between-system variation that discrimination requires an account consistent with our data, though we do not directly isolate the contrastive-training variable, which we leave to future work.
\item We give a concrete \textbf{reporting recipe} place every metric in (validity, discriminative-power) space, with an explicit random floor and along the way show that raw BERTScore F1 (85.3--86.6\%) overstates alignment by 1--3 points over random, that baseline rescaling exposes only a 6--13-point signal, and that response length explains little of the real-vs-random gap under per-reply scoring (median $R^2 \approx 3\%$).
\item We validate the findings with an \textbf{LLM-as-judge} protocol: three independent judges from three developers reproduce the validity axis with large effect sizes (paired Cohen's $d = 2.0$-$3.5$) and high inter-judge agreement (Krippendorff's $\alpha \ge 0.85$); like the automatic metrics, the judges separate the five models only weakly.
\end{enumerate}

\section{Methodology}
We study how to \emph{reliably measure} the alignment of LLM-generated answers with community answers on open-ended questions, organized around three research questions about what an evaluation metric must do. \textbf{RQ1 (validity):} Once each metric is paired with an empirical random-derangement baseline, how much of its score reflects genuine content alignment rather than surface coincidence? \textbf{RQ2 (discriminative power):} How well does each metric distinguish one system from another? \textbf{RQ3 (the tradeoff):} Do validity and discriminative power trade off across metrics, and what in the metrics' design produces that tradeoff? We additionally verify that the validity signal is not an artifact of response length and corroborate it with three independent LLM judges.

\begin{table}[t]
\centering
\small
\renewcommand{\arraystretch}{1.5}
\resizebox{\columnwidth}{!}{
\begin{tabular}{lcc}
\toprule
\textbf{Model} & \textbf{Training Cutoff} & \textbf{Verification Source} \\
\midrule
Mistral-7B-v0.3 & $\leq$ May 2024\textsuperscript{$\dagger$} & \href{https://huggingface.co/mistralai/Mistral-7B-Instruct-v0.3}{HuggingFace metadata} \\
Qwen2.5-7B-Inst. & $\leq$ Dec 2023\textsuperscript{$\ddagger$} & \href{https://github.com/QwenLM/Qwen2.5/issues/525}{Developer disclosure on GitHub} \\
Granite3.1-Dense-8B & Apr 2024\textsuperscript{$\dagger$}  & \href{https://ollama.com/library/granite3.1-dense:8b}{Ollama - Model Release Date} \\
Falcon3-10B-Inst. & $\leq$ Dec 2024\textsuperscript{$\dagger$} & \href{https://huggingface.co/blog/falcon3}{HuggingFace Falcon Blog} \\
Llama-3.1-8B-Inst. & Dec 2023 & \href{https://huggingface.co/meta-llama/Llama-3.1-8B-Instruct}{HuggingFace Model Card} \\
\bottomrule
\end{tabular}
}
\caption{Training data cutoffs for evaluated models. All cutoffs 
predate the RECOM data collection window (September 2025) by at 
least 9 months, ruling out test set contamination. 
\textsuperscript{$\dagger$}Cutoff not officially disclosed; model 
release date used as a conservative upper bound. 
\textsuperscript{$\ddagger$}Stated by the Qwen team on GitHub.}
\label{tab:cutoffs}
\end{table}

\subsection{Dataset}
We introduce RECOM (Reddit Evaluation for Correspondence of Models), an evaluation dataset for assessing LLM alignment with community perspectives on open-ended questions. We collected 132,728 English-language posts from r/AskReddit (September 1--30, 2025) using the Pushshift API. From these, we retained the top 25,000 by engagement (total comments and replies) and randomly sampled 15,000 to form RECOM; after the refusal filtering described in Section~\ref{sec:filtering}, all reported results use the resulting 11{,}528 questions.

For each question, the evaluation reference is the set of depth-1 community replies the direct human answers posted in the thread. Using the authentic replies rather than a synthesized reference preserves the full distribution of community perspectives, including minority viewpoints, and avoids artifacts from an intermediate generation step. These replies should be interpreted as community-aligned human answers rather than authoritative ground truth, since a popular opinion need not be correct. Most r/AskReddit questions do not require temporally recent factual knowledge; the September 2025 collection window ensures all community replies postdate the training cutoffs of every evaluated model, preventing data contamination, as Table~\ref{tab:cutoffs} verifies.

\subsection{Models and Inference}

We evaluate five open-source LLMs spanning 7B--10B parameters (Table~\ref{tab:model-details}), all served through Ollama. Each model was prompted with identical instructions (Table~\ref{tab:ans-gen}) constraining responses to under 50 words. Inference parameters (Table~\ref{tab:inference-config}) were held constant across all models (temperature=0.8, top-$p$=0.9, top-$k$=40, seed=42). All inference was conducted on four NVIDIA RTX A5000 GPUs (24GB each).

\begin{table}[t]
\centering
\small
\begin{tabular}{lcc}
\toprule
\textbf{Model} & \textbf{Params} & \textbf{Disk Size} \\
\midrule
Mistral-7B & 7.2B & 4.4 GB \\
Qwen2.5-7B & 7.6B & 4.7 GB \\
Granite3.1-Dense-8B & 8.2B & 5.0 GB \\
Falcon3-10B & 10.3B & 6.3 GB \\
Llama-3.1-8B & 8.0B & 4.9 GB \\
\bottomrule
\end{tabular}
\caption{Evaluated models. All are dense transformers with BF16 original precision, served via Ollama with uniform Q4\_0 quantization.}
\label{tab:model-details}
\end{table}

\subsection{Response Filtering}\label{sec:filtering}

We filtered responses containing AI self-references (``as an AI,'' ``I'm a language model''), capability disclaimers (``I don't have,'' ``I cannot experience''), and role disclaimers (``I'm not human''). Questions where \emph{any} model produced a refusal were removed, yielding 11,528 question-response pairs per model (76.9\% retention).

\subsection{Evaluation Metrics}\label{sec:metrics}

We use a multi-reference evaluation framework: we treat all $N$ depth-1 replies for a post as the reference set the authentic human answers to the question and compute every metric for each (reply, generated answer) pair. Per-post scores are then aggregated across replies, with the choice of aggregation matched to each metric family's literature convention: \textbf{max} across references for BLEU \cite{papineni2002bleu}, ROUGE \cite{lin2004rouge}, and BERTScore \cite{zhang2019bertscore} (the standard multi-reference protocol a candidate need only match one reference well), and \textbf{mean} across references for cosine similarity, MoverScore, and natural language inference (NLI), where each reply contributes equally to a community-level alignment score. Both the max and the mean aggregation are computed for every metric; the results tables and the validity--discrimination figure report the family-convention choice above, while the length-confound regression and the per-metric effect-size heatmap report results for both. Across our $11{,}528$ posts the reference set has a mean of $56.2$ replies and a median of $14$, giving $647{,}973$ total (reply, answer) comparisons per model. To establish the noise floor for each metric, we additionally compute every metric under a \emph{random-derangement baseline} where each post is paired with a different post's generated answer (fixed seed; no fixed points), holding length and surface style fixed while breaking content correspondence.

\textbf{Lexical:} Cumulative BLEU-1 through BLEU-4 \cite{papineni2002bleu} and ROUGE-1, ROUGE-2, ROUGE-L F1 \cite{lin2004rouge} with Porter stemming. BLEU is scored per (reply, answer) pair with Method~1 smoothing, as sentence-level BLEU requires rather than at the canonical corpus level, because our per-post effect-size analysis (Cohen's $d$, paired tests, and the random-derangement gap) requires a score for every individual post.

\textbf{Semantic:} BERTScore precision, recall, and F1 using RoBERTa-large (layer 12) \cite{zhang2019bertscore,liu2019roberta}, reported both raw and baseline-rescaled; MoverScore via Earth Mover's Distance over DistilBERT embeddings \cite{zhao2019moverscore}; and cosine similarity using contrastive sentence-transformers (\texttt{all-mpnet-base-v2}, \texttt{all-MiniLM-L6-v2}) \cite{reimers2019sentence}.

\textbf{Inference-based:} NLI classification using BART-large-MNLI \cite{lewis2020bart,williams2018multinli}. For each (reply, generated answer) pair we split the generated answer into sentences and classify each sentence against the reply as entailment, contradiction, or neutral; the per-pair score is the proportion of sentences in each class. Per-post NLI is the mean of these proportions across all replies, matching our multi-reference aggregation policy for inference-based metrics.

\textbf{Statistical tests:} Wilcoxon signed-rank test \cite{casella2024statistical} for pairwise comparisons, with Cohen's $d$ effect sizes to assess practical significance at $n = 11{,}528$.

These metric families capture different aspects of alignment. Lexical metrics emphasize surface overlap, semantic metrics capture embedding-based similarity, and inference-based metrics assess logical consistency. We therefore interpret results collectively rather than relying on any single metric.

\subsection{LLM-as-Judge Protocol}\label{sec:judge}

The automatic metrics above could in principle register a real-vs-random gap for reasons unrelated to genuine community alignment, such as surface fluency, topical vocabulary, or residual length effects. To test whether the gap reflects substantive alignment, we additionally evaluate with an LLM-as-judge protocol, which prior work has shown correlates well with human judgment \citep{kocmi2023large}.

\textbf{Judges.} We use three instruction-tuned LLMs as independent judges: Phi-4 (14B, Microsoft), Gemma-3-27B-Instruct (Google), and Yi-1.5-34B-Chat (01.AI). The three come from three different developers and share no lineage with the five evaluated models, avoiding self-preference bias. All judges are served locally through Ollama at temperature~0 with JSON-constrained output for deterministic, parseable scoring.

\textbf{Sample and conditions.} From the 11{,}528 posts we draw a reproducible random sample of 5{,}000 (NumPy \texttt{RandomState(42)}, without replacement) and query the three judges over it. Scoring three local judge models was compute-bound, so the analysis uses a fixed \textbf{1{,}000-post subsample} of this seed-42 sample, which we verify is statistically indistinguishable from the rest of the draw (Appendix~\ref{app:judge-data}). Each judge scores all five models' answers to each of the 1{,}000 posts under both the \emph{real} and \emph{random-derangement} conditions of Section~\ref{sec:metrics} ten judgments per post so the judge results are directly comparable, post for post, to the automatic-metric real-vs-random gaps.

\textbf{Dimensions.} Each judge rates two dimensions on a 1--5 Likert scale. \emph{Relevancy} measures how directly the generated answer addresses the question; the judge sees only the question and the answer. \emph{Faithfulness} measures how well the answer reflects the perspectives expressed in the community discussion; the judge additionally sees up to 50 chronologically-earliest depth-1 replies. Since the median post has 14 replies, most sampled posts are scored against their entire discussion. The two prompt templates are given in Appendix~\ref{app:judge-prompts}.

\textbf{Agreement.} To quantify reliability we report Krippendorff's ordinal $\alpha$ \citep{krippendorff2011computing} across the three judges, computed on the 10{,}000 (post, model, condition) items each judge scored, together with pairwise quadratically-weighted Cohen's $\kappa$.

\section{Results}
We evaluated five open-source LLMs (Mistral-7B, Qwen2.5-7B, Granite3.1-Dense-8B, Falcon3-10B, and Llama-3.1-8B) on 11{,}528 Reddit questions after filtering refusal responses. Each model's generated answer was scored against every depth-1 reply for the corresponding post (multi-reference protocol of Section~\ref{sec:metrics}) and against the random-derangement baseline. Given the large sample size, even small differences are statistically significant under Wilcoxon signed-rank tests; we therefore report Cohen's $d$ effect sizes throughout and treat statistical significance as a sanity check rather than the primary criterion. For every metric we report two quantities its \emph{validity} (the real-vs-random effect size) and its \emph{discriminative power} (the largest inter-model effect size, computed between real model outputs) and Section~\ref{sec:tradeoff} places all metrics on these two axes.

\subsection{Lexical Overlap Metrics}
Table~\ref{tab:lexical_overlap} presents BLEU and ROUGE scores under the multi-reference max protocol, alongside random-derangement baselines. The five models cluster tightly: BLEU-1 spans only 21.09\%--22.63\% (a 1.07$\times$ spread across the five models) and ROUGE-L F1 spans 16.72\%--21.78\%. This compression and the substantially higher absolute scores reflect the multi-reference setup with a median of 14 and a mean of 56 depth-1 replies per post, every model can match at least one reply well, so max-aggregation lifts the lexical floor. The random baseline yields BLEU-1 of 15.55\%--17.55\% and ROUGE-L F1 of 11.72\%--12.84\%, establishing a 5--9 percentage-point gap that all real generations clear consistently. Lexical metrics under this protocol distinguish content match from random pairing reliably (real $>$ random for every model on every BLEU/ROUGE variant) but do not strongly differentiate models from each other.

\begin{table*}[!tb]
\centering
\small
\renewcommand{\arraystretch}{1.5}
\begin{tabular}{lccccc}
\toprule
\textbf{Model} & \textbf{BLEU-1} & \textbf{BLEU-4} & \textbf{ROUGE-1 F1} & \textbf{ROUGE-2 F1} & \textbf{ROUGE-L F1} \\
\midrule
Mistral-7B    & 21.67 $\pm$ 7.46  & 2.69 $\pm$ 3.48 & 23.28 $\pm$ 8.54  & 6.16 $\pm$ 7.02  & 18.08 $\pm$ 7.75  \\
Qwen2.5-7B    & 22.10 $\pm$ 11.73 & 3.78 $\pm$ 5.23 & 23.73 $\pm$ 15.01 & 6.37 $\pm$ 11.39 & 21.09 $\pm$ 14.56 \\
Granite3.1-Dense-8B & 21.09 $\pm$ 6.95  & 2.26 $\pm$ 2.69 & 22.18 $\pm$ 7.59  & 5.37 $\pm$ 5.52  & 16.72 $\pm$ 6.30  \\
Falcon3-10B     & 22.63 $\pm$ 11.56 & 3.73 $\pm$ 5.38 & 25.09 $\pm$ 13.86 & 7.21 $\pm$ 11.21 & 21.78 $\pm$ 13.62 \\
Llama-3.1-8B    & 22.04 $\pm$ 7.64  & 2.50 $\pm$ 3.33 & 23.38 $\pm$ 8.67  & 5.72 $\pm$ 6.42  & 17.38 $\pm$ 7.60  \\
\midrule
\textit{Random baseline (avg)} & \textit{16.56} & \textit{1.52} & \textit{15.24} & \textit{1.50} & \textit{12.27} \\
\bottomrule
\end{tabular}
\caption{Lexical Overlap Metrics under the multi-reference \emph{max} aggregation (mean $\pm$ std across 11{,}528 posts, in percentages). Real generations cluster tightly between 21.09\%--22.63\% on BLEU-1; the random-derangement baseline (last row, mean across the five models) sits at 16.56\%, giving a $\sim$5--9 pp real-vs-random gap. Per-model random baselines fall within $\pm 1$ pp of the mean. With $n = 11{,}528$, 95\% confidence intervals are narrow and are omitted for brevity.}
\label{tab:lexical_overlap}
\end{table*}

\begin{table}[t]
\centering
\small
\renewcommand{\arraystretch}{1.5}
\resizebox{0.80\columnwidth}{!}{
\begin{tabular}{lc}
\toprule
\textbf{Model} & \textbf{Mean Word Count} \\
\midrule
Mistral-7B    & 27.3 \\
Qwen2.5-7B    & 11.9 \\
Granite3.1-Dense-8B & 32.4 \\
Falcon3-10B     & 15.1 \\
Llama-3.1-8B    & 32.8 \\
\midrule
\textit{Depth-1 replies (reference set, avg)} & \textit{19.1} \\
\bottomrule
\end{tabular}
}
\caption{Mean response word counts per model alongside the average length of a depth-1 reply (our multi-reference unit). Despite identical ``under 50 words'' prompts, models produce responses ranging from 11.9 to 32.8 words a 2.8$\times$ difference. Mistral, Granite, and Llama ($\sim$27--33 words) sit close to or slightly above the average reply length (19.1 words), while Qwen and Falcon3 (12--15 words) are systematically below it.}
\label{tab:response-length}
\end{table}

\begin{table*}[!tb]
\centering
\small
\renewcommand{\arraystretch}{1.5}
\resizebox{\textwidth}{!}{
\begin{tabular}{lccccccccc}
\toprule
\textbf{Model} & \textbf{BS P} & \textbf{BS R} & \textbf{BS F1} & \textbf{BS P (Resc.)} & \textbf{BS R (Resc.)} & \textbf{BS F1 (Resc.)} & \textbf{MVS} & \textbf{Cos (MPNet)} & \textbf{Cos (MiniLM)} \\
\midrule
Mistral-7B            & 85.24 & 88.14 & 85.59 & 19.11 & 34.99 & 21.11 & $-$1.61 & 28.63 & 25.31 \\
Qwen2.5-7B            & 86.70 & 88.66 & 86.59 & 27.08 & 37.87 & 26.62 & $-$0.76 & 28.81 & 25.01 \\
Granite3.1-Dense-8B   & 84.83 & 88.06 & 85.34 & 16.88 & 34.55 & 19.78 & $-$2.08 & 28.71 & 25.14 \\
Falcon3-10B           & 86.39 & 88.61 & 86.37 & 25.41 & 37.55 & 25.40 & $-$1.03 & 28.98 & 25.33 \\
Llama-3.1-8B          & 84.79 & 88.46 & 85.44 & 16.65 & 36.73 & 20.33 & $-$1.51 & 29.20 & 25.19 \\
\midrule
\textit{Random baseline (avg)} & \textit{83.49} & \textit{87.04} & \textit{84.25} & \textit{9.53} & \textit{28.96} & \textit{13.83} & \textit{$-$6.73} & \textit{4.69} & \textit{5.46} \\
\bottomrule
\end{tabular}
}
\caption{Semantic Similarity Metrics under multi-reference aggregation (mean, in percentages). BS = BERTScore (RoBERTa-large, layer 12), MVS = MoverScore, Cos = Cosine Similarity. Resc.\ = baseline-rescaled \cite{zhang2019bertscore}. BERTScore uses \textbf{max} across all depth-1 replies (literature convention); MoverScore and Cosine use \textbf{mean} across replies (community-level alignment). The random-derangement baseline (last row, mean across the five models) makes the calibration story explicit: raw BERTScore F1 sits 1--3 pp above random, rescaled BERTScore F1 sits 6--13 pp above, and Cosine (MPNet) sits $\sim$24 pp above ($\sim$28.6\% vs 4.7\%). MoverScore is unbounded below (formulated as $1 - \mathrm{EMD}$); negative real-run means reflect that the average individual reply is far from the candidate in DistilBERT embedding space, but real generations remain $\sim$5 pp above the random baseline of 6.7. With $n = 11{,}528$, 95\% confidence intervals are narrow and are omitted for brevity.}
\label{tab:semantic-similarity}
\end{table*}

Multi-reference model rankings are also unstable across metric variants. On BLEU-1 every pairwise Cohen's $d$ is negligible ($|d| < 0.14$); on ROUGE-L F1 the largest gap is between Granite3.1-Dense-8B and Falcon3-10B ($d = -0.40$), and other pairs remain in the small-effect range. The ranking also reshuffles by metric variant: Falcon3-10B leads on BLEU-1 (22.63\%) and ROUGE-L F1 (21.78\%), Qwen2.5-7B leads on BLEU-4 (3.78\%) and ROUGE-2 F1 (6.37\%), and Granite3.1-Dense-8B trails on most variants. Parameter count carries no obvious signal either: Falcon3-10B (10B) and Llama-3.1-8B (8B) sit within 0.6 pp of Mistral-7B (7B) on BLEU-1. Under multi-reference evaluation, lexical metrics primarily certify that a model produces fluent English overlapping with the community vocabulary they distinguish real from random generations well, but do not strongly discriminate among real models.

\subsection{Response Length Analysis}

Response length varies substantially across models despite the identical ``under 50 words'' prompt constraint (Table~\ref{tab:response-length}). Llama-3.1-8B (mean 32.8 words) and Granite3.1-Dense-8B (32.4) produce the longest responses, Mistral-7B (27.3) sits in the middle, and Qwen2.5-7B (11.9) and Falcon3-10B (15.1) are noticeably terser. Depth-1 replies that serve as our reference set are themselves short (median 17 words, mean 19), closely matching the candidate lengths, so the candidate-vs-reference length asymmetry is modest.

Response length explains little of the real-vs-random gap. Because the random-derangement baseline is a permutation of the same generated answers, the candidate-length distribution is identical in the real and random conditions, so the gap cannot be a marginal-length artifact by construction. We additionally verify this per-post: for each metric we compute the real-minus-random difference and regress that gap on the candidate's and the reference set's word counts, across all 21 metrics $\times$ 2 aggregation variants $\times$ 5 models ($n = 210$ fits, each on 11{,}528 posts). If the gap were a verbosity artifact, longer responses would show systematically larger gaps. They do not: the median $R^2$ is $0.034$, 85\% of fits fall below $R^2 = 0.10$, and only Length\_Ratio (which \emph{is} length, by construction) exceeds $R^2 = 0.25$. For BLEU-1 the gap regression gives a median $R^2$ of $1.4\%$, for BERTScore F1 $7.0\%$, BERTScore precision $10.1\%$, MoverScore $1.6\%$, and Cosine (MPNet) $4.1\%$; the raw metric scores are themselves only weakly length-dependent (median $R^2 = 0.073$ over the same fits). The real-vs-random gaps in Tables~\ref{tab:lexical_overlap} and~\ref{tab:semantic-similarity} therefore reflect content alignment rather than verbosity artifacts.

\subsection{Semantic Similarity}
Semantic metrics under multi-reference aggregation show substantially larger \emph{raw} inter-model spread than lexical ones, and the calibration story is sharper. Raw BERTScore F1 (max across references) is numerically high (85.3\%--86.6\%) but uninformative without baseline calibration: rescaled BERTScore F1 spans 19.8\%--26.6\%, against a random-derangement floor of 13.8\%, leaving a 6--13 pp signal of genuine content alignment. Cosine similarity (mean across replies, MPNet) clusters tightly at 28.6\%--29.2\% versus a random baseline of 4.7\% a $\sim$24 pp gap, the largest of any metric we evaluate. MoverScore computed as $1 - \mathrm{EMD}$ (Earth Mover's Distance) over DistilBERT embeddings and unbounded below has real-run means of $2.1$\% to $0.8$\% versus a random baseline of $6.7$\%, giving a $\sim$5--6 pp gap; the negative magnitudes reflect that a typical individual reply is far from the candidate in embedding space (community responses cover many perspectives the candidate does not), but the consistent real-vs-random delta confirms genuine content signal.

Under multi-reference aggregation, BERTScore precision has the highest \emph{raw} discriminative power (inter-model separation) of any metric we evaluate, with BERTScore F1 a close second both open a substantially wider raw inter-model spread than the lexical or cosine metrics. For precision, the largest pairwise effect sizes are Qwen2.5-7B vs.\ Llama-3.1-8B and Qwen2.5-7B vs.\ Granite3.1-Dense-8B, each reaching Cohen's $d \approx 0.63$ (medium), with Granite3.1-Dense-8B vs.\ Falcon3-10B at $d = -0.58$; the implied precision ordering is Qwen2.5-7B $>$ Falcon3-10B $>$ Mistral-7B $>$ Granite3.1-Dense-8B $\approx$ Llama-3.1-8B. F1 separates models slightly less Qwen2.5-7B vs.\ Granite3.1-Dense-8B reaches $d = 0.53$ (medium) and Falcon3-10B vs.\ Granite3.1-Dense-8B reaches $d = -0.46$, with an F1 ordering of Qwen2.5-7B $\approx$ Falcon3-10B $>$ Mistral-7B $\approx$ Llama-3.1-8B $>$ Granite3.1-Dense-8B. The calibration concern remains essential for both raw precision (84.8\%--86.7\%) and raw F1 (85.3\%--86.6\%) sit only 1--3 pp above their random floors, while the rescaled values (precision 16.7\%--27.1\%, F1 19.8\%--26.6\%) open gaps of 7--18 pp and 6--13 pp above random. Max-aggregation under per-reply scoring rewards matching \emph{some} reply well, which favors concise candidates that can closely echo a short reply.

\textbf{Discriminative power is largely a length effect.} This concision preference has a direct consequence for the discrimination axis. The BERTScore-precision model ordering Qwen2.5-7B $>$ Falcon3-10B $>$ Mistral-7B $>$ Granite3.1-Dense-8B $\approx$ Llama-3.1-8B is, to within ties, the five models sorted from shortest to longest mean response (Table~\ref{tab:response-length}). Because precision rewards candidates whose tokens are fully covered by a short reference, much of what we report as discriminative power reflects sensitivity to response length rather than to answer quality. Our length analysis controls the real-vs-random gap; extending the same control to the inter-model comparison confirms the concern directly. Residualizing each candidate's own word count out of the metric a single pooled regression across all five models' outputs (Appendix~\ref{app:residualization}) collapses the maximum inter-model $|d|$ from 0.63 to 0.09 for BERTScore precision and from 0.53 to 0.11 for BERTScore F1: once response length is controlled, both fall to a negligible level, on par with cosine similarity. MoverScore is a partial exception: its smaller raw separation ($|d| \le 0.37$) is barely changed by the same control ($|d| = 0.34$ residualized; length $R^2 \approx 0$), so it reflects genuine, length-independent system variation though MoverScore's own validity is only modest. Genuine discriminative power thus survives length control, but it is small and confined to a metric that does not also lead on validity. We therefore read the raw discrimination axis of Figure~\ref{fig:tradeoff} as a length-inflated upper bound, and take the length-residualized effect size as the estimate of genuine, quality-driven system separation.

Cosine similarity shows the opposite pattern from BERTScore: high \emph{validity} (a huge real-vs-random separation) but minimal \emph{discriminative power}. MPNet cosine clusters at 28.6\%--29.2\% across all five models (a 0.6 pp range), MiniLM at 25.0\%--25.3\% (0.3 pp). The random-derangement baseline sits at 4.7\% (MPNet) and 5.5\% (MiniLM), giving real-vs-random gaps of 23--24 pp on MPNet and 19--20 pp on MiniLM the largest real-vs-random separation of any metric we evaluate. All pairwise model effect sizes are negligible ($|d| < 0.1$ on MPNet), and the ranking is essentially flat across both embedding backbones. Cosine similarity is therefore a strong content-vs-noise indicator but a poor model-ranking tool under this multi-reference protocol.

MoverScore sits between the two patterns: real-run means of $2.1$\% to $0.8$\% versus a random baseline of $6.7$\% give a $\sim$5--6 pp content signal, modest relative to cosine's 24 pp gap. Inter-model discrimination is moderate ($|d| \le 0.37$), with Qwen2.5-7B vs.\ Granite3.1-Dense-8B at $d = 0.37$ being the largest pair. The model ordering Qwen2.5-7B $>$ Falcon3-10B $>$ Mistral-7B $\approx$ Llama-3.1-8B $>$ Granite3.1-Dense-8B coincides with the raw BERTScore F1 ordering rather than the (flat) cosine one. The two are not, however, driven by a shared signal: the length residualization above collapses BERTScore F1's separation (to $|d| = 0.11$) but leaves MoverScore's intact ($|d| = 0.34$), so BERTScore's ordering is largely a length artifact while MoverScore's is not.

\textbf{Precision-Recall Patterns:} Under max-aggregation, all five models show recall \emph{above} precision (e.g., Mistral-7B: $P = 85.2\%$, $R = 88.1\%$; Granite3.1-Dense-8B: $P = 84.8\%$, $R = 88.1\%$), with rescaled recall sitting 11--20 pp above rescaled precision across all models. This reflects the multi-reference geometry: a typical depth-1 reply (19 words) is close in length to the candidate (12--33 words), so the best-matching reply per post often has most of its tokens covered by the candidate, lifting recall. Every model sits $\sim$6--9 pp above the random rescaled-recall floor (29.0\%), so no model falls below chance in coverage.

\section{Conclusion}

We introduced RECOM, a contamination-free dataset of 15{,}000 open-ended r/AskReddit questions, and used it to ask what automatic metrics actually deliver when evaluating LLMs on opinion-driven question answering. Our central finding is a \emph{validity--discrimination tradeoff}: across the lexical, semantic, and inference-based metrics we evaluate, none both separates genuine content alignment from surface coincidence (\emph{validity}) and reliably ranks one system above another (\emph{discriminative power}). Cosine similarity maximizes validity but cannot tell the five models apart; BERTScore precision appears to rank them best, but that apparent ranking is largely a response-length effect ($|d|$ from 0.63 to 0.09 once length is controlled), and its own validity margin over noise is modest. Because every metric scores the identical model outputs, this is a property of the metrics, not the models one we attribute to representation design: the contrastive geometry that lets an encoder sharply separate related from unrelated text appears to be the same geometry that collapses the between-system variation discriminative power needs. Three independent LLM judges reproduce the validity axis, confirming the signal is not an artifact of any single metric family; like the automatic metrics, they too separate the five models only weakly.

The practical consequence is that ``does this output engage with the content?'' and ``which model is better?'' are different questions which, on open-ended tasks, no single metric answers; metrics should therefore be reported on both axes, with an explicit random-baseline floor. The tradeoff follows from metric design rather than from our task; whether it recurs in other open-ended settings summarization, dialogue, long-form QA is a natural question for future work, as our evidence is limited to short-form opinion QA. We release RECOM to support this line of evaluation research. 

\section{Limitations}

In the LLM-as-judge validation, one of the three judges (Yi-1.5-34B) shows a mild positive correlation between its Relevancy scores and response length ($\rho \approx 0.19$), a length sensitivity not present in the other two judges; we therefore report the three judges' scores separately rather than pooling them. The LLM-as-judge validation is run on a fixed 1{,}000-post subsample rather than the full dataset, and the judges' scores are not yet calibrated against human ratings.

Our dataset is limited to English-language r/AskReddit posts (September 1--30, 2025), a subreddit emphasizing opinion-based questions. RECOM is moreover drawn from the most-engaged posts of the collection window (the top 25{,}000 by comment and reply count), so it over-represents high-traffic threads rather than uniformly sampling r/AskReddit. Findings may not generalize to factual QA or domains with objectively verifiable answers. All models use 4-bit quantization (Q4\_0), which may affect quality differently across architectures. We evaluate a single prompting strategy without retrieval augmentation; performance may differ under alternative prompts or decoding strategies. Our metrics measure alignment with community consensus rather than factual correctness high similarity could mask agreement with incorrect responses. Finally, the multi-reference protocol introduces a stylistic-heterogeneity concern: the reference set is a heterogeneous mix of individual community replies (jokes, anecdotes, brief opinions, longer arguments), so each candidate is compared against a target set with varying registers and intentions. Max- and mean-aggregation across replies only partially smooth over this heterogeneity. Two further limitations bear on our central axes. First, the discriminative-power axis is largely a response-length effect: our metrics' model rankings track mean answer length, and residualizing candidate length collapses the largest inter-model $|d|$ for BERTScore precision from 0.63 to 0.09. The discriminative-power magnitudes are also specific to the narrow 7--10B quality band we evaluate: on models of more widely differing quality the absolute inter-model effect sizes could shift, and it is the \emph{relative} validity--discrimination pattern, not these magnitudes, that we take to be robust. Because we do not establish an independent quality ranking of the five models, a small inter-model effect size admits two readings a metric insensitive to genuine quality differences, or models that are themselves close in quality. That three independent LLM judges also separate the five models only weakly (Section~\ref{sec:judge-results}) is consistent with the latter; cleanly disentangling the two would require models with a known, larger quality gap. Second, the validity axis establishes that a metric separates genuine content correspondence from topical coincidence, but not that it tracks answer \emph{quality}; quantifying quality-tracking would require human ratings, which we leave to future work.

\section{Ethical Considerations}

This work analyzes publicly available Reddit data. We do not deanonymize users or release user identifiers. The community replies that serve as reference answers reflect the views of users active in each thread and may over-represent dominant viewpoints; they should not be interpreted as authoritative ground truth. All LLMs are used strictly for research purposes, and findings should not be interpreted as guarantees of correctness for real-world applications. RECOM is released for research use only (CC~BY-NC~4.0): the release contains question text, depth-1 reply text, and Reddit post IDs, with all usernames removed. r/AskReddit is predominantly safe-for-work; we did not perform explicit NSFW or toxicity screening, which we note as a limitation of the release.

\bibliography{custom}

@inproceedings{papineni2002bleu,
  title={Bleu: a method for automatic evaluation of machine translation},
  author={Papineni, Kishore and Roukos, Salim and Ward, Todd and Zhu, Wei-Jing},
  booktitle={Proceedings of the 40th annual meeting of the Association for Computational Linguistics},
  pages={311--318},
  year={2002}
}

@inproceedings{lin2004rouge,
  title={Rouge: A package for automatic evaluation of summaries},
  author={Lin, Chin-Yew},
  booktitle={Text summarization branches out},
  pages={74--81},
  year={2004}
}

@article{zhang2019bertscore,
  title={Bertscore: Evaluating text generation with bert},
  author={Zhang, Tianyi and Kishore, Varsha and Wu, Felix and Weinberger, Kilian Q and Artzi, Yoav},
  journal={arXiv preprint arXiv:1904.09675},
  year={2019}
}

@article{liu2019roberta,
  title={Roberta: A robustly optimized bert pretraining approach},
  author={Liu, Yinhan and Ott, Myle and Goyal, Naman and Du, Jingfei and Joshi, Mandar and Chen, Danqi and Levy, Omer and Lewis, Mike and Zettlemoyer, Luke and Stoyanov, Veselin},
  journal={arXiv preprint arXiv:1907.11692},
  year={2019}
}

@inproceedings{lewis2020bart,
  title={BART: Denoising sequence-to-sequence pre-training for natural language generation, translation, and comprehension},
  author={Lewis, Mike and Liu, Yinhan and Goyal, Naman and Ghazvininejad, Marjan and Mohamed, Abdelrahman and Levy, Omer and Stoyanov, Veselin and Zettlemoyer, Luke},
  booktitle={Proceedings of the 58th annual meeting of the association for computational linguistics},
  pages={7871--7880},
  year={2020}
}

@book{casella2024statistical,
  title={Statistical inference},
  author={Casella, George and Berger, Roger},
  year={2024},
  publisher={Chapman and Hall/CRC}
}

@inproceedings{zhao2019moverscore,
  title     = {MoverScore: Text Generation Evaluating with Contextualized Embeddings and Earth Mover Distance},
  author    = {Zhao, Wei and Peyrard, Maxime and Liu, Fei and Gao, Yang and Meyer, Christian M. and Eger, Steffen},
  booktitle = {Proceedings of the 2019 Conference on Empirical Methods in Natural Language Processing and the 9th International Joint Conference on Natural Language Processing (EMNLP-IJCNLP)},
  year      = {2019},
  pages     = {563--578}
}

@inproceedings{reimers2019sentence,
  title     = {Sentence-BERT: Sentence Embeddings using Siamese BERT-Networks},
  author    = {Reimers, Nils and Gurevych, Iryna},
  booktitle = {Proceedings of the 2019 Conference on Empirical Methods in Natural Language Processing and the 9th International Joint Conference on Natural Language Processing (EMNLP-IJCNLP)},
  year      = {2019},
  pages     = {3982--3992}
}

@article{feinstein1990high,
  title={High Agreement but Low Kappa: {I}. The Problems of Two Paradoxes},
  author={Feinstein, Alvan R and Cicchetti, Domenic V},
  journal={Journal of Clinical Epidemiology},
  volume={43},
  number={6},
  pages={543--549},
  year={1990}
}

@article{vu2023freshllms,
  title={Fresh{LLM}s: Refreshing Large Language Models with Search Engine Augmentation},
  author={Vu, Tu and Iyyer, Mohit and Wang, Xuezhi and Constant, Noah and Wei, Jerry and Wei, Jason and Tar, Chris and Sung, Yun-Hsuan and Zhou, Denny and Le, Quoc and Luong, Thang},
  journal={arXiv preprint arXiv:2310.03214},
  year={2023}
}

@inproceedings{kasai2024realtime,
  title={{RealTime QA}: What's the Answer Right Now?},
  author={Kasai, Jungo and Sakaguchi, Keisuke and Takahashi, Yoichi and Le Bras, Ronan and Asai, Akari and Yu, Xinyan and Radev, Dragomir and Smith, Noah A. and Choi, Yejin and Inui, Kentaro},
  booktitle={Advances in Neural Information Processing Systems 36 (NeurIPS Datasets and Benchmarks Track)},
  year={2023}
}

@inproceedings{chen2021timeqa,
  title={A Dataset for Answering Time-Sensitive Questions},
  author={Chen, Wenhu and Wang, Xinyi and Wang, William Yang},
  booktitle={Proceedings of NeurIPS Datasets and Benchmarks},
  year={2021}
}

@inproceedings{kocmi2023large,
  title={Large language models are state-of-the-art evaluators of translation quality},
  author={Kocmi, Tom and Federmann, Christian},
  booktitle={Proceedings of the 24th Annual Conference of the European Association for Machine Translation},
  pages={193--203},
  year={2023}
}

@techreport{krippendorff2011computing,
  title={Computing {K}rippendorff's Alpha-Reliability},
  author={Krippendorff, Klaus},
  year={2011},
  institution={University of Pennsylvania, Annenberg School for Communication}
}

@inproceedings{zheng2023judging,
  title={Judging {LLM}-as-a-Judge with {MT}-Bench and Chatbot Arena},
  author={Zheng, Lianmin and Chiang, Wei-Lin and Sheng, Ying and Zhuang, Siyuan and Wu, Zhanghao and Zhuang, Yonghao and Lin, Zi and Li, Zhuohan and Li, Dacheng and Xing, Eric P. and Zhang, Hao and Gonzalez, Joseph E. and Stoica, Ion},
  booktitle={Advances in Neural Information Processing Systems (NeurIPS)},
  year={2023}
}

@inproceedings{liu2023geval,
  title={{G-Eval}: {NLG} Evaluation using {GPT-4} with Better Human Alignment},
  author={Liu, Yang and Iter, Dan and Xu, Yichong and Wang, Shuohang and Xu, Ruochen and Zhu, Chenguang},
  booktitle={Proceedings of the 2023 Conference on Empirical Methods in Natural Language Processing (EMNLP)},
  year={2023}
}

@inproceedings{williams2018multinli,
  title={A Broad-Coverage Challenge Corpus for Sentence Understanding through Inference},
  author={Williams, Adina and Nangia, Nikita and Bowman, Samuel R.},
  booktitle={Proceedings of the 2018 Conference of the North American Chapter of the Association for Computational Linguistics: Human Language Technologies},
  pages={1112--1122},
  year={2018}
}

\appendix

\section{Appendix}
\label{sec:appendix}

\subsection{Computational Budget}
\label{app:compute}

All experiments used NVIDIA RTX A5000 (24\,GB) GPUs on a shared SLURM cluster. Answer generation across the five models ran on four A5000 GPUs. Automatic-metric computation BERTScore, MoverScore, cosine similarity, and NLI, for both the real and random-derangement conditions across all five models ran one metric per GPU and totalled approximately 250--300 GPU-hours, with NLI (BART-large-MNLI) the dominant cost. The three LLM judges were each served on a single A5000 via Ollama; producing the 1{,}000-post analysis set took one to three days of wall-clock per judge.

\subsection{Length-Residualized Discrimination}
\label{app:residualization}

The length-residualized inter-model effect sizes reported in Section~\ref{sec:tradeoff} are computed as follows. For each metric we pool all five models' per-post scores and candidate word counts ($5 \times 11{,}528$ observations) and fit a single ordinary least-squares regression of the metric score on candidate word count. A single pooled slope rather than a per-model regression, which would absorb each model's mean and trivially zero all inter-model differences removes the shared length effect while preserving genuine between-model variation. We then take residuals (observed minus fitted score) and recompute the inter-model paired Cohen's $d_z$ on them exactly as for the raw scores. Only candidate word count is residualized: reference-set length is constant across models within a post and so cannot contribute to inter-model differences.

\subsection{LLM-as-Judge Data Preparation}
\label{app:judge-data}

The LLM-as-judge validation (Section~\ref{sec:judge}) operates on a random subsample of the 11{,}528-post evaluation set. We first draw a reproducible random sample of 5{,}000 posts (\texttt{np.random.RandomState(42).choice}, without replacement); for efficient I/O the sampled indices are sorted into ascending parquet-row order, which because the table is keyed on the ascending Reddit post ID is effectively chronological. The three judges process the resulting rows (5 models $\times$ 2 conditions per post) strictly sequentially; scoring three local judge models was compute-bound, so all three completed at least the first 1{,}087 posts of this ordering, and the analysis (Section~\ref{sec:judge-results}) uses a fixed 1{,}000-post target, set in advance. Because the 5{,}000 posts were randomly sampled \emph{before} being sorted, the chronological ordering does not bias the analysis set: the 1{,}000 analysis posts are statistically indistinguishable from the remaining posts of the random draw in depth-1 reply count and generated-answer length (mean 23.7 vs.\ 23.9 words). Table~\ref{tab:judge-data} summarizes the pipeline.

\begin{table}[t]
\centering
\small
\renewcommand{\arraystretch}{1.3}
\resizebox{\columnwidth}{!}{
\begin{tabular}{llcccc}
\toprule
\textbf{Dimension} & \textbf{Judge} & \textbf{Real} & \textbf{Random} & \textbf{Gap} & \textbf{Cohen's }$d$ \\
\midrule
\multirow{3}{*}{Relevancy} & Phi-4 & 4.36 & 1.21 & 3.15 & 3.35 \\
\cmidrule{2-6}
 & Gemma-3 & 4.59 & 1.21 & 3.38 & 3.48 \\
\cmidrule{2-6}
 & Yi-1.5 & 4.07 & 1.20 & 2.88 & 3.21 \\
\midrule
\multirow{3}{*}{Faithfulness} & Phi-4 & 3.20 & 1.11 & 2.09 & 2.42 \\
\cmidrule{2-6}
 & Gemma-3 & 3.18 & 1.17 & 2.00 & 2.03 \\
\cmidrule{2-6}
 & Yi-1.5 & 2.85 & 1.18 & 1.67 & 2.05 \\
\bottomrule
\end{tabular}
}
\caption{LLM-as-judge scores (1--5 Likert) under the real and random-derangement conditions, on a fixed 1{,}000-post subsample ($n = 5{,}000$ paired (post, model) observations per row). All three judges register a large real-vs-random gap (paired Cohen's $d = 2.0$-$3.5$) on both dimensions; every gap is significant at $p < 0.001$.}
\label{tab:judge-realrandom}
\end{table}

\begin{table}[t]
\centering
\small
\renewcommand{\arraystretch}{1.3}
\begin{tabular}{lcccc}
\toprule
\textbf{Dimension} & \textbf{Kripp. }$\bm{\alpha}$ & \textbf{P\,$\times$\,G} & \textbf{P\,$\times$\,Y} & \textbf{G\,$\times$\,Y} \\
\midrule
Relevancy  & 0.898 & 0.940 & 0.929 & 0.906 \\
Faithfulness & 0.853 & 0.880 & 0.824 & 0.796 \\
\bottomrule
\end{tabular}
\caption{Inter-judge agreement on the 10{,}000 (post, model, condition) items scored by all three judges. Krippendorff's $\alpha$ is the ordinal three-judge reliability; P\,$\times$\,G, P\,$\times$\,Y, G\,$\times$\,Y are pairwise quadratically-weighted Cohen's $\kappa$ between Phi-4 (P), Gemma-3 (G), and Yi-1.5 (Y). Both dimensions exceed the $\alpha \ge 0.80$ reliability threshold.}
\label{tab:judge-agreement}
\end{table}

\begin{table}[t]
\centering
\small
\renewcommand{\arraystretch}{1.3}
\begin{tabular}{lr}
\toprule
\textbf{Stage} & \textbf{Count} \\
\midrule
Evaluation set (posts)        & 11{,}528 \\
Random judge sample, seed 42 (posts)    & 5{,}000 \\
Analysis subsample (posts)        & 1{,}000 \\
Judgments per judge ($1{,}000\times5\times2$) & 10{,}000 \\
\bottomrule
\end{tabular}
\caption{LLM-as-judge data pipeline. The 1{,}000-post analysis subsample is a fixed block of the seed-42 random sample, verified statistically indistinguishable from the rest of the draw.}
\label{tab:judge-data}
\end{table}

\subsection{LLM-as-Judge Prompts}
\label{app:judge-prompts}

Tables~\ref{tab:judge-rel-prompt} and~\ref{tab:judge-faith-prompt} give the prompt templates used for the LLM-as-judge validation (Section~\ref{sec:judge}). Each judge is queried once per dimension; the placeholders \texttt{question}, \texttt{gen\_answer}, and \texttt{replies} are filled per item. All three judges are run at temperature~0 with JSON-constrained output.

\begin{table}[t]
  \centering
  \small
  \renewcommand{\arraystretch}{1.4}
  \begin{tabular}{p{0.89\columnwidth}}
  \toprule
  \textbf{Relevancy prompt (question + generated answer).} \\
  \midrule
  You are evaluating a generated answer to a Reddit question. \newline
  \newline
  QUESTION: \texttt{question} \newline
  GENERATED ANSWER: \texttt{gen\_answer} \newline
  \newline
  Rate the generated answer on RELEVANCY how directly and completely it addresses the question on a 1--5 scale (1 = off-topic; 2 = tangentially related; 3 = partially addresses; 4 = mostly addresses; 5 = directly and completely addresses the question). Respond only with valid JSON containing a one-sentence \texttt{reasoning} field and an integer \texttt{relevancy} field (1--5). \\
  \bottomrule
  \end{tabular}
  \caption{LLM-as-judge Relevancy prompt.}
  \label{tab:judge-rel-prompt}
\end{table}

\begin{table}[t]
  \centering
  \small
  \renewcommand{\arraystretch}{1.4}
  \begin{tabular}{p{0.89\columnwidth}}
  \toprule
  \textbf{Faithfulness prompt (question + generated answer + community replies).} \\
  \midrule
  You are evaluating a generated answer's alignment with community perspectives on a Reddit question. \newline
  \newline
  QUESTION: \texttt{question} \newline
  GENERATED ANSWER: \texttt{gen\_answer} \newline
  COMMUNITY REPLIES (up to 50 chronologically-earliest depth-1 replies): \texttt{replies} \newline
  \newline
  Rate the generated answer on FAITHFULNESS how well it reflects the perspectives expressed in the community replies on a 1--5 scale (1 = conflicts with or ignores them; 2 = mostly diverges; 3 = partially overlaps; 4 = mostly reflects; 5 = closely reflects the dominant community perspectives). Respond only with valid JSON containing a one-sentence \texttt{reasoning} field and an integer \texttt{faithfulness} field (1--5). \\
  \bottomrule
  \end{tabular}
  \caption{LLM-as-judge Faithfulness prompt.}
  \label{tab:judge-faith-prompt}
\end{table}

\subsection{Sample Questions and Model Responses}
\label{sec:appendix-dataset}

Table~\ref{tab:dataset-sample} shows two sample questions and the corresponding response from each of the five evaluated models, illustrating the diversity of open-ended r/AskReddit questions and the range of answer styles models produce under the ``under 50 words'' prompt.

\begin{table*}[t]
\centering
\small
\setlength{\tabcolsep}{4pt}
\renewcommand{\arraystretch}{1.3}
\begin{tabular}{l p{0.38\textwidth} p{0.40\textwidth}}
\toprule
\textbf{Model} & \textbf{RECOM Question} & \textbf{Model Response} \\
\midrule
\multirow{2}{*}{Mistral-7B} & You're in charge of renaming Earth. What's the new name? & Globus Terra Prime \\
\cmidrule{2-3}
 & What's the best pop tart flavour for someone who has never had one? & The classic Strawberry Pop-Tart is a popular choice for first-timers, offering a balance of sweetness and fruity flavor. However, personal preference plays a role; other options include Brown Sugar Cinnamon or S'mores, depending on taste preferences. \\
\midrule
\multirow{2}{*}{Qwen2.5-7B} & What app/site were you in right before using Reddit? & I was on Facebook. \\
\cmidrule{2-3}
 & Does money actually buy happiness? If yes, in what way? & Money can buy some things that contribute to happiness, like basic needs and experiences, but true long-term happiness comes from other sources like relationships and personal growth. \\
\midrule
\multirow{2}{*}{Granite3.1-Dense-8B} & What is the tv show most recognizable by one line? & ``The Simpsons'' - ``D'oh!'' \\
\cmidrule{2-3}
 & What's one small thing that makes life just THAT much better / easier? & A well-organized closet or pantry can significantly simplify daily routines, making life more enjoyable. Efficient storage reduces decision fatigue and saves time searching for items. \\
\midrule
\multirow{2}{*}{Falcon3-10B} & what's the first thing you do when you wake up? & Brushing my teeth. \\
\cmidrule{2-3}
 & If there were ever a `Museum of Internet History', what permanent/temporary exhibits should it have? & Permanent exhibits: The creation of ARPANET, the birth of the World Wide Web, the rise of social media platforms. Temporary exhibits: Historic hacks and cyber incidents, Evolution of web design trends, Milestones in internet accessibility worldwide. \\
\midrule
\multirow{2}{*}{Llama-3.1-8B} & Which game did you play for hundreds of hours but never finished? & World of Warcraft. \\
\cmidrule{2-3}
 & What is the least stressful job? & Park Ranger, as they often work outdoors in natural environments, have regular schedules, and interact with animals rather than people. This combination can lead to a calming and peaceful work atmosphere. \\
\bottomrule
\end{tabular}
\caption{Two sample questions and the corresponding response from each of the five evaluated models, illustrating the diversity of open-ended r/AskReddit questions and the range of answer styles from one-line answers to short multi-sentence responses produced under the ``under 50 words'' prompt.}
\label{tab:dataset-sample}
\end{table*}

\subsection{NLI Classification Results}
\label{app:nli}

Table~\ref{tab:nli-results} reports NLI classification results under the multi-reference protocol. Neutral classifications account for 60\%--76\% of sentence-reply pairs, contradiction for 22\%--36\%, and entailment for 1.9\%--3.6\%. \emph{Entailment} provides the cleanest content signal a 4--7$\times$ multiplicative gap over the random-derangement baseline (0.48\%) though absolute rates remain modest, indicating that direct logical support from any individual reply is uncommon. \emph{Contradiction} rates are inconsistent across models and show no clean real-vs-random signal: Qwen2.5-7B (35.9\%) and Falcon3-10B (34.3\%) sit roughly 10--12 pp above the random baseline of 23.9\%, Llama-3.1-8B about 4 pp above, and Mistral-7B (21.8\%) and Granite3.1-Dense-8B (23.8\%) fall at or just below it. Because BART-large-MNLI is applied out of domain on these Reddit (reply, answer) pairs, this contradiction signal should be read with caution.

\begin{table}[t]
  \centering
  \small
  \renewcommand{\arraystretch}{1.5}
  \resizebox{0.99\columnwidth}{!}{
  \begin{tabular}{lccc}
  \toprule
  \textbf{Model} & \textbf{Entailment} & \textbf{Contradiction} & \textbf{Neutral} \\
  \midrule
  Mistral-7B    & 2.15\% & 21.80\% & 76.05\% \\
  Qwen2.5-7B    & 3.64\% & 35.87\% & 60.49\% \\
  Granite3.1-Dense-8B & 2.27\% & 23.75\% & 73.98\% \\
  Falcon3-10B     & 3.16\% & 34.30\% & 62.54\% \\
  Llama-3.1-8B    & 1.90\% & 27.82\% & 70.27\% \\
  \midrule
  \textit{Random baseline (avg)} & \textit{0.48\%} & \textit{23.94\%} & \textit{75.58\%} \\
\bottomrule
\end{tabular}
}
\caption{NLI classification distribution (BART-large-MNLI) under the multi-reference protocol: each candidate sentence is classified against each individual depth-1 reply, then averaged across all replies per post. Real generations produce 1.9\%--3.6\% entailment versus a random-derangement baseline of 0.48\% (a 4--7$\times$ increase); contradiction rates (21.8\%--35.9\%) are inconsistent, ranging from just below the random baseline (23.9\%) to $\sim$12 pp above it.}
\label{tab:nli-results}
\end{table}

\subsection{The Validity--Discrimination Tradeoff}\label{sec:tradeoff}

\begin{figure}[t]
\centering
\includegraphics[width=\columnwidth]{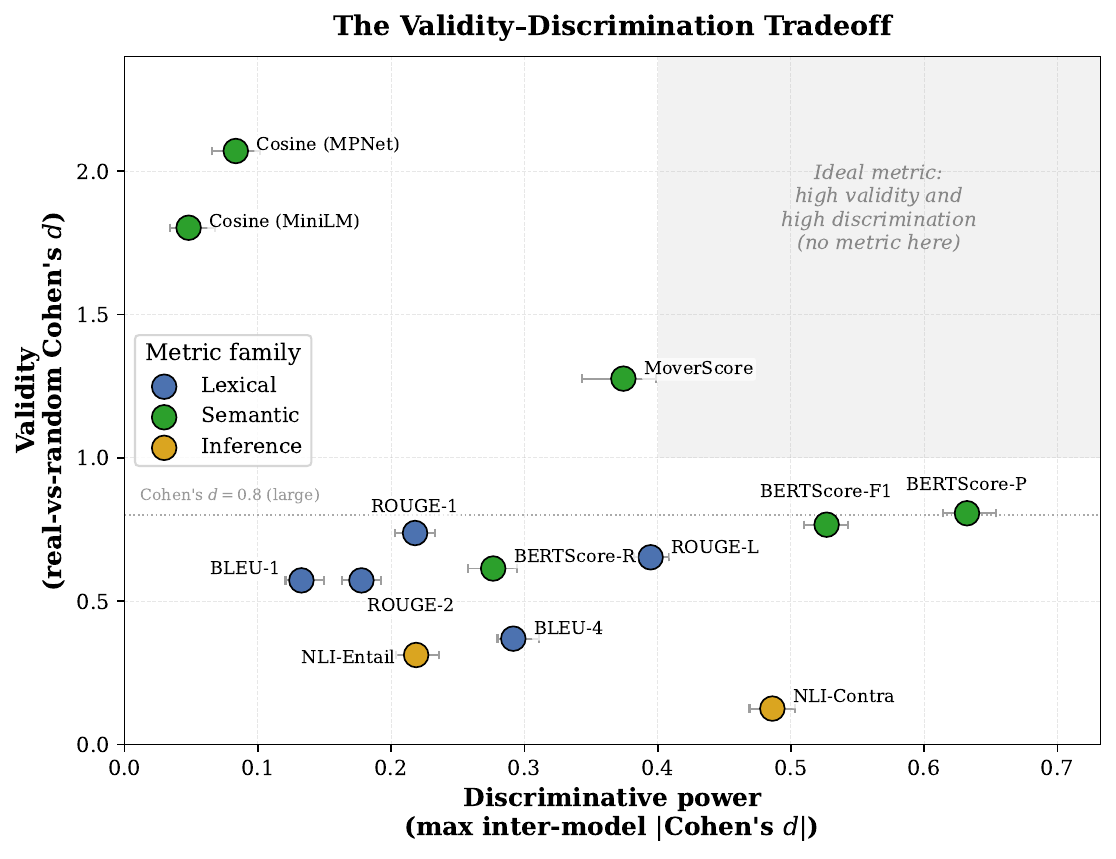}
\caption{The validity--discrimination tradeoff. Each metric is a point: the $x$axis is \emph{discriminative power} (the largest inter-model effect size, $\max|d|$) and the $y$axis is \emph{validity} (the real-vs-random effect size, Cohen's $d$); colors denote metric family. Error bars are paired bootstrap 95\% confidence intervals (1{,}000 resamples over the 11{,}528 posts); they are narrow at this sample size and the BERTScore-precision and BERTScore-F1 intervals on the discrimination axis do not overlap. No metric reaches the top-right region (shaded), where an ideal metric high on both axes would sit: cosine similarity has the highest validity ($d \approx 2$) but almost no discriminative power, while the strongest \emph{raw} model discriminators (BERTScore precision and F1) sit far lower on validity. The discrimination axis is raw inter-model $|d|$; residualizing response length collapses it (BERTScore precision $0.63 \to 0.09$), so this axis is best read as a length-inflated upper bound.}
\label{fig:tradeoff}
\end{figure}

A metric is asked to do two things: separate genuine content alignment from surface coincidence its \emph{validity}, which we measure as the real-vs-random effect size and separate one system from another its \emph{discriminative power}, the largest inter-model effect size. Validity in this sense is a \emph{necessary} but not a sufficient property: clearing the random floor shows a metric responds to topical content correspondence, not that it tracks answer quality, which our design does not test directly. Figure~\ref{fig:tradeoff} plots every metric on these two axes: no metric occupies the high-validity, high-discrimination corner, and across the metrics we evaluate higher validity tends to accompany lower discriminative power. We refer to this pattern as the \emph{validity--discrimination tradeoff}, noting that with a modest number of metric points it is a qualitative pattern rather than a precisely estimated relationship. It is, however, robust to the multi-reference aggregation choice: recomputing validity and discrimination for every metric under both \texttt{max}- and \texttt{mean}-aggregation leaves the high-validity, high-discrimination corner empty under either protocol, with cosine similarity high on validity and low on discrimination both ways. Cosine similarity has the highest validity the largest real-vs-random separation ($\sim$24~pp on MPNet, $d \approx 2$) but essentially no discriminative power ($|d| < 0.1$), and produces no meaningful model ranking. BERTScore precision is the reverse: the strongest \emph{raw} discriminative power ($|d|$ up to 0.63, though residualizing length cuts this to 0.09) but markedly weaker validity ($d \approx 0.8$, versus cosine's $\approx 2$) once baseline-rescaled, it opens only a 7--18~pp margin over its random floor. Lexical metrics (BLEU, ROUGE) and MoverScore sit in the middle on both raw axes though, as the length analysis above shows, MoverScore is the one mid-axis metric whose discrimination survives length control and the inference-based NLI metrics sit low on validity (Appendix~\ref{app:nli}). Model rankings are correspondingly inconsistent across families Qwen2.5-7B leads on BERTScore F1, BLEU-4, ROUGE-2, and MoverScore but trails on raw lexical coverage so the answer to ``which model is best?'' depends on which metric one asks. No single metric answers both questions at once.

\textbf{Why the tradeoff arises.} The tradeoff is not a coincidence of two hand-picked metrics; we argue it follows from how the metrics represent text. Cosine similarity is computed over \emph{contrastively-trained} sentence encoders (MPNet, MiniLM), whose objective explicitly pulls paraphrases and topically related sentences together and pushes unrelated text apart. That geometry is what gives cosine its validity on-topic answers and random pairings land far apart but it is also what destroys its discriminative power: because the objective collapses all topically related text into a tight neighborhood, five different on-topic answers to the same question receive almost the same score, and cosine's per-model means span only 0.6~pp on MPNet (0.3~pp on MiniLM). BERTScore behaves oppositely. Its token-level RoBERTa embeddings are not contrastively trained for sentence similarity, so absolute scores are compressed near the top (85.3\%--86.6\%, hence the need for rescaling) weak raw validity but greedy token matching preserves fine-grained between-model variation: rescaled BERTScore precision spans 10.4~pp across the five models, an order of magnitude wider than cosine's range though, as shown above, this between-model spread is itself largely a response-length effect rather than a quality signal. The representational property that buys validity a geometry that sharply separates related from unrelated text is therefore the very property that erases the between-system variation a discriminator would need. Because every metric scores the identical set of model outputs, the tradeoff is a property of the metrics, not the models. Directly isolating the contrastive-training variable for instance with a non-contrastive encoder baseline would further test this account and is left to future work.

\subsection{LLM-as-Judge Results}\label{sec:judge-results}

We now turn from the automatic metrics to the LLM-as-judge protocol described in Section~\ref{sec:judge}.

\textbf{Judges reproduce the real-vs-random gap.} Table~\ref{tab:judge-realrandom} reports the judge scores. All three judges register a large real-vs-random gap on both dimensions: paired Cohen's $d$ ranges from 3.21 to 3.48 for Relevancy and from 2.03 to 2.42 for Faithfulness, far above the $d \ge 0.8$ threshold for a ``large'' effect. Real generations average 4.07--4.59 of 5 on Relevancy versus 1.20--1.21 for random pairings, and 2.85--3.20 versus 1.11--1.18 on Faithfulness; every gap is significant at $p < 0.001$ (paired $t > 143$; $n = 5{,}000$, i.e.\ $1{,}000$ posts $\times$ 5 models). Because the judges share no lineage with the evaluated models and were given no automatic-metric scores, this independently confirms that the real-vs-random gap is not an artifact specific to any single metric family.

\textbf{Judges agree with one another.} Inter-judge reliability is high (Table~\ref{tab:judge-agreement}): Krippendorff's ordinal $\alpha$ is 0.898 for Relevancy and 0.853 for Faithfulness, both above the 0.80 threshold conventionally taken to indicate reliable measurement, and pairwise quadratically-weighted $\kappa$ ranges from 0.80 to 0.94. Agreement is slightly lower for Faithfulness than Relevancy, consistent with Faithfulness being the more subjective judgment. These figures are computed over the pooled real and random-derangement items; within either condition alone the 1--5 scores are range-restricted, where Krippendorff's $\alpha$ is deflated even under high raw agreement \citep{feinstein1990high}, so the pooled estimate is the more interpretable measure of reliability.

\textbf{Judges separate the five models only weakly.} The discriminative-power axis asks whether an evaluator can rank one system above another. Computing inter-model paired Cohen's $d$ for the judges (real condition, same 1{,}000-post sample) shows that they largely cannot: on Faithfulness the largest inter-model effect size is negligible for every judge ($\max|d| = 0.10$-$0.16$), and on Relevancy, where the separation is larger ($\max|d| = 0.40$-$0.57$), the model ordering tracks mean response length longer-answering models are rated more relevant with the effect largest for Yi-1.5, the one judge with a documented length sensitivity (Limitations). An independent, judgment-based evaluator thus reaches the automatic metrics' verdict: the five 7--10B models are hard to separate on quality, and the separation that does surface is entangled with length rather than content.

\textbf{Convergent validity with automatic metrics.} The judge scores correlate with the automatic metrics in the expected direction: judge Faithfulness correlates most strongly with embedding cosine similarity (Spearman $\rho$ up to 0.42), as both measure semantic alignment with the community replies. The agreement of an independent, judgment-based evaluator with the embedding-based metrics provides convergent validity for the multi-reference protocol. Combined with prior evidence that LLM judges track human judgments \citep{kocmi2023large, liu2023geval, zheng2023judging}, the automatic-metric findings rest on two anchors an established, human-correlated judging method and the agreement of three independent-vendor judges though direct human calibration of the judges on RECOM remains future work.

\subsection{Implications for Metric Usage}

Our results carry three practical implications for researchers evaluating LLMs on open-ended tasks.

\textbf{BERTScore requires baseline calibration.} Raw BERTScore should not be reported without either rescaling or an empirical baseline floor: in our multi-reference setting, raw F1 ranges from 85.3\% to 86.6\% numerically high but only 1--3~pp above the random-derangement baseline (84.2\%). Rescaled values (19.8\%--26.6\%, against a rescaled-random floor of 13.8\%) make the actual content signal visible at 6--13~pp above random. A raw BERTScore F1 of 85\% does not mean ``85\% similar to the reference'' in any absolute sense. We recommend reporting both rescaled scores and an explicit random-baseline column whenever BERTScore is the headline metric.

\textbf{Length confounds depend on reference structure.} When references are much longer than candidate responses, brevity-penalty effects can dominate lexical metrics independently of content. Under our multi-reference per-reply scoring, references are short and length-matched to candidates (mean 19 words), and an ordinary least-squares (OLS) regression of the per-post real-vs-random gap on candidate and reference word counts yields a median $R^2$ of just $1.4\%$ for BLEU-1, with similarly small values for BERTScore F1 (7.0\%), BERTScore precision (10.1\%), MoverScore (1.6\%), and cosine MPNet (4.1\%) length explains little of the gap. The practical recommendation: evaluations should report the length distribution of the reference set and the candidate-to-reference length ratio. When that ratio exceeds 2$\times$, length-residualized scores should be reported alongside raw values; when references are length-matched to candidates, as in our protocol, the lexical signal is more directly interpretable as content overlap.

\textbf{Report validity and discriminative power separately.} Model rankings differ across metric families under multi-reference evaluation a direct consequence of the validity--discrimination tradeoff (Section~\ref{sec:tradeoff}), not a flaw of multi-metric reporting. Cosine similarity has high \emph{validity} but almost no \emph{discriminative power} (real-vs-random $|d| > 2$, inter-model $|d| < 0.1$); BERTScore precision is the reverse the strongest \emph{raw} discriminative power ($|d|$ up to 0.63, though largely a length effect) but a modest validity margin once its raw 84\% floor is calibrated away; lexical metrics validate content overlap but cluster models tightly. Reporting any single metric invites mischaracterizing a model: the metric that certifies ``this output is genuine content'' and the metric that certifies ``model A beats model B'' are, on open-ended QA, not the same metric. We therefore recommend multi-metric evaluation in which every metric is reported on both axes, with an explicit random-baseline floor, so readers can separately answer ``does this model produce content-matching text?'' and ``how does it compare to others?'' Because the discrimination axis can itself absorb response-length differences between systems (Section~\ref{sec:tradeoff}), inter-model effect sizes should additionally be reported length-residualized, so that apparent system separation is not a verbosity artifact.

\textbf{Independent LLM judges corroborate the real-vs-random signal.} A natural concern is that the automatic-metric gaps reflect surface artifacts rather than content alignment. Our LLM-as-judge validation (Section~\ref{sec:judge-results}) addresses this directly: three judges from three different developers, given no automatic-metric scores, all reproduce the real-vs-random gap with large effect sizes (paired Cohen's $d = 2.0$-$3.5$) and agree strongly with one another (Krippendorff's $\alpha \ge 0.85$). Convergent evidence from automatic metrics and from independent judges makes the headline real-vs-random finding robust to the choice of evaluator.

\subsection{Distribution Visualizations}
\label{app:figures}

\begin{figure}[t]
\centering
\includegraphics[width=0.92\columnwidth]{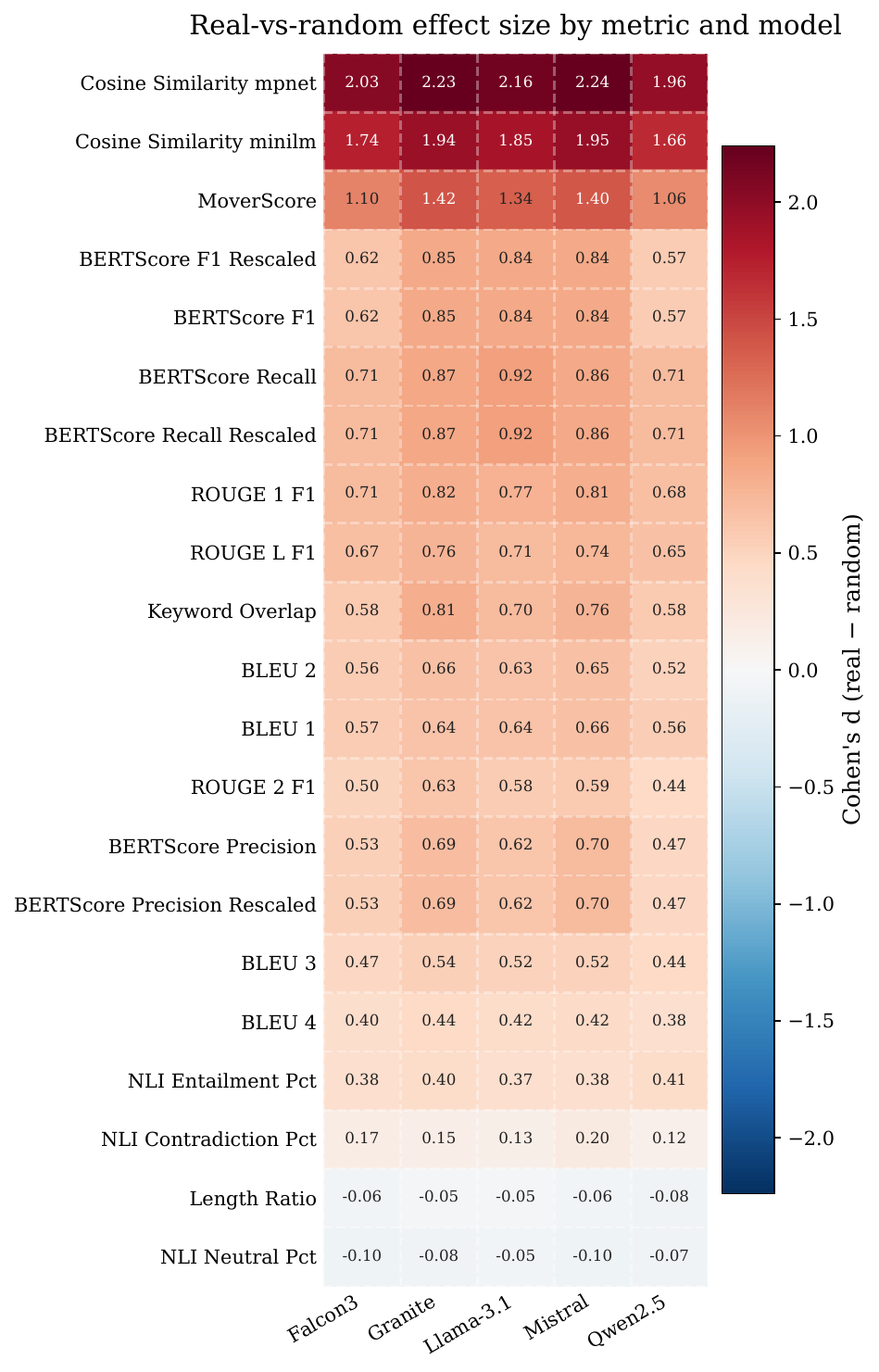}
\caption{Cohen's $d$ for the real-vs-random difference, per metric and model, averaged across the two aggregation variants (\texttt{all\_max} and \texttt{all\_mean}). Cosine similarity shows the largest effect ($d \approx 2$), followed by MoverScore and BERTScore; Length\_Ratio and NLI\_Neutral show essentially no real-vs-random separation.}
\label{fig:cohensd-heatmap}
\end{figure}

This section reports the automatic-metric results in detail, contrasting the \emph{real} condition (generated answer vs.\ community replies) with the \emph{random-derangement} baseline. Figure~\ref{fig:cohensd-heatmap} reports the real-vs-random effect size per metric and model. Tables~\ref{tab:stats-bleu}--\ref{tab:stats-length_quality} then give per-metric descriptive statistics mean, median, standard deviation, minimum, maximum, and mode for the \texttt{all\_max} and \texttt{all\_mean} aggregation variants, with the random-baseline mean reported alongside each real-condition mean.

\begin{table}[t]
  \centering
  \small
  \renewcommand{\arraystretch}{1.5}
  \begin{tabular}{p{0.80\columnwidth}}
  \toprule
  \textbf{Answer Generation Prompt} \\
  \midrule
  You are answering this question.
  Give a direct answer without
  restating or summarizing the question. \newline
  Rules: \newline
  - Answer ONLY \newline
  - Be clear and concise \newline
  - Keep it under 50 words \\
  \bottomrule
  \end{tabular}
  \caption{Answer Generation Prompt}
  \label{tab:ans-gen}
\end{table}

\begin{table}[t]
\centering
\small
\renewcommand{\arraystretch}{1.5}
\begin{tabular}{ll}
\toprule
\textbf{Parameter} & \textbf{Value} \\
\midrule
Serving framework & Ollama 0.12.3 \\
Context window (\texttt{num\_ctx}) & 8,192 tokens \\
Max generation (\texttt{num\_predict}) & 512 tokens \\
Temperature & 0.8 \\
Top-$p$ (nucleus) & 0.9 \\
Top-$k$ & 40 \\
Seed & 42 \\
\bottomrule
\end{tabular}
\caption{Inference configuration for all evaluated models.}
\label{tab:inference-config}
\end{table}

\begin{table*}[t]
\centering
\footnotesize
\setlength{\tabcolsep}{4pt}
\resizebox{\textwidth}{!}{
\begin{tabular}{lllrrrrrrr}
\toprule
\textbf{Metric} & \textbf{Model} & \textbf{Variant} & \textbf{Mean (real)} & \textbf{Mean (rand.)} & \textbf{Median} & \textbf{Std} & \textbf{Min} & \textbf{Max} & \textbf{Mode} \\
\midrule
\multirow{10}{*}{BLEU\_1} & \multirow{2}{*}{Mistral} & all\_max & 0.22 & 0.17 & 0.21 & 0.075 & 0 & 1.00 & 0.20 \\
 &  & all\_mean & 0.074 & 0.056 & 0.072 & 0.030 & 0 & 0.30 & 0 \\
\cmidrule{2-10}
 & \multirow{2}{*}{Qwen2.5} & all\_max & 0.22 & 0.16 & 0.20 & 0.12 & 0 & 1.00 & 0.25 \\
 &  & all\_mean & 0.061 & 0.042 & 0.057 & 0.031 & 0 & 0.32 & 0 \\
\cmidrule{2-10}
 & \multirow{2}{*}{Granite} & all\_max & 0.21 & 0.17 & 0.21 & 0.069 & 0 & 0.67 & 0.20 \\
 &  & all\_mean & 0.073 & 0.057 & 0.071 & 0.030 & 0 & 0.26 & 0.062 \\
\cmidrule{2-10}
 & \multirow{2}{*}{Falcon3} & all\_max & 0.23 & 0.16 & 0.21 & 0.12 & 0 & 1.00 & 0.25 \\
 &  & all\_mean & 0.067 & 0.046 & 0.065 & 0.032 & 0 & 0.34 & 0 \\
\cmidrule{2-10}
 & \multirow{2}{*}{Llama-3.1} & all\_max & 0.22 & 0.18 & 0.22 & 0.076 & 0 & 1.00 & 0.20 \\
 &  & all\_mean & 0.076 & 0.059 & 0.073 & 0.031 & 0 & 0.34 & 0.045 \\
\midrule
\multirow{10}{*}{BLEU\_2} & \multirow{2}{*}{Mistral} & all\_max & 0.086 & 0.046 & 0.075 & 0.061 & 0 & 1.00 & 0.031 \\
 &  & all\_mean & 0.020 & 0.012 & 0.017 & 0.012 & 0 & 0.25 & 0 \\
\cmidrule{2-10}
 & \multirow{2}{*}{Qwen2.5} & all\_max & 0.092 & 0.045 & 0.058 & 0.098 & 0 & 1.00 & 0.058 \\
 &  & all\_mean & 0.019 & 0.011 & 0.016 & 0.013 & 0 & 0.21 & 0 \\
\cmidrule{2-10}
 & \multirow{2}{*}{Granite} & all\_max & 0.078 & 0.044 & 0.072 & 0.051 & 0 & 0.59 & 0.080 \\
 &  & all\_mean & 0.019 & 0.012 & 0.016 & 0.011 & 0 & 0.21 & 0 \\
\cmidrule{2-10}
 & \multirow{2}{*}{Falcon3} & all\_max & 0.096 & 0.045 & 0.071 & 0.093 & 0 & 1.00 & 0 \\
 &  & all\_mean & 0.020 & 0.012 & 0.017 & 0.014 & 0 & 0.27 & 0 \\
\cmidrule{2-10}
 & \multirow{2}{*}{Llama-3.1} & all\_max & 0.085 & 0.050 & 0.075 & 0.060 & 0 & 1.00 & 0.075 \\
 &  & all\_mean & 0.020 & 0.013 & 0.018 & 0.012 & 0 & 0.23 & 0 \\
\midrule
\multirow{10}{*}{BLEU\_3} & \multirow{2}{*}{Mistral} & all\_max & 0.045 & 0.021 & 0.028 & 0.047 & 0 & 1.00 & 0.019 \\
 &  & all\_mean & 0.011 & 0.007 & 0.009 & 0.007 & 0 & 0.22 & 0 \\
\cmidrule{2-10}
 & \multirow{2}{*}{Qwen2.5} & all\_max & 0.054 & 0.027 & 0.032 & 0.070 & 0 & 1.00 & 0.022 \\
 &  & all\_mean & 0.012 & 0.007 & 0.010 & 0.008 & 0 & 0.17 & 0 \\
\cmidrule{2-10}
 & \multirow{2}{*}{Granite} & all\_max & 0.039 & 0.019 & 0.026 & 0.038 & 0 & 0.57 & 0.029 \\
 &  & all\_mean & 0.010 & 0.007 & 0.008 & 0.005 & 0 & 0.18 & 0 \\
\cmidrule{2-10}
 & \multirow{2}{*}{Falcon3} & all\_max & 0.055 & 0.025 & 0.032 & 0.070 & 0 & 1.00 & 0 \\
 &  & all\_mean & 0.012 & 0.007 & 0.010 & 0.009 & 0 & 0.24 & 0 \\
\cmidrule{2-10}
 & \multirow{2}{*}{Llama-3.1} & all\_max & 0.043 & 0.021 & 0.027 & 0.045 & 0 & 1.00 & 0.024 \\
 &  & all\_mean & 0.010 & 0.007 & 0.009 & 0.006 & 0 & 0.16 & 0 \\
\midrule
\multirow{10}{*}{BLEU\_4} & \multirow{2}{*}{Mistral} & all\_max & 0.027 & 0.013 & 0.016 & 0.035 & 0 & 0.78 & 0.016 \\
 &  & all\_mean & 0.007 & 0.005 & 0.006 & 0.004 & 0 & 0.19 & 0 \\
\cmidrule{2-10}
 & \multirow{2}{*}{Qwen2.5} & all\_max & 0.038 & 0.021 & 0.023 & 0.052 & 0 & 1.00 & 0.022 \\
 &  & all\_mean & 0.009 & 0.006 & 0.008 & 0.006 & 0 & 0.13 & 0 \\
\cmidrule{2-10}
 & \multirow{2}{*}{Granite} & all\_max & 0.023 & 0.012 & 0.015 & 0.027 & 0 & 0.56 & 0.017 \\
 &  & all\_mean & 0.006 & 0.005 & 0.006 & 0.003 & 0 & 0.14 & 0 \\
\cmidrule{2-10}
 & \multirow{2}{*}{Falcon3} & all\_max & 0.037 & 0.018 & 0.022 & 0.054 & 0 & 1.00 & 0.022 \\
 &  & all\_mean & 0.009 & 0.006 & 0.008 & 0.006 & 0 & 0.20 & 0 \\
\cmidrule{2-10}
 & \multirow{2}{*}{Llama-3.1} & all\_max & 0.025 & 0.013 & 0.015 & 0.033 & 0 & 1.00 & 0.015 \\
 &  & all\_mean & 0.007 & 0.005 & 0.006 & 0.004 & 0 & 0.11 & 0 \\
\bottomrule
\end{tabular}}
\caption{BLEU metrics: descriptive statistics across the 11{,}528 posts. Mean, median, std, min, max, and mode are for the \emph{real} condition; Mean (rand.) is the random-derangement baseline. Variants: \texttt{all\_max} and \texttt{all\_mean} aggregation over depth-1 replies.}
\label{tab:stats-bleu}
\end{table*}

\begin{table*}[t]
\centering
\footnotesize
\setlength{\tabcolsep}{4pt}
\resizebox{\textwidth}{!}{
\begin{tabular}{lllrrrrrrr}
\toprule
\textbf{Metric} & \textbf{Model} & \textbf{Variant} & \textbf{Mean (real)} & \textbf{Mean (rand.)} & \textbf{Median} & \textbf{Std} & \textbf{Min} & \textbf{Max} & \textbf{Mode} \\
\midrule
\multirow{10}{*}{ROUGE\_1\_F1} & \multirow{2}{*}{Mistral} & all\_max & 0.23 & 0.16 & 0.22 & 0.085 & 0 & 1.00 & 0.25 \\
 &  & all\_mean & 0.083 & 0.053 & 0.079 & 0.039 & 0 & 0.38 & 0 \\
\cmidrule{2-10}
 & \multirow{2}{*}{Qwen2.5} & all\_max & 0.24 & 0.13 & 0.21 & 0.15 & 0 & 1.00 & 0 \\
 &  & all\_mean & 0.065 & 0.033 & 0.060 & 0.043 & 0 & 0.41 & 0 \\
\cmidrule{2-10}
 & \multirow{2}{*}{Granite} & all\_max & 0.22 & 0.16 & 0.22 & 0.076 & 0 & 1.00 & 0.25 \\
 &  & all\_mean & 0.081 & 0.054 & 0.078 & 0.037 & 0 & 0.36 & 0 \\
\cmidrule{2-10}
 & \multirow{2}{*}{Falcon3} & all\_max & 0.25 & 0.15 & 0.23 & 0.14 & 0 & 1.00 & 0.25 \\
 &  & all\_mean & 0.076 & 0.042 & 0.071 & 0.043 & 0 & 0.40 & 0 \\
\cmidrule{2-10}
 & \multirow{2}{*}{Llama-3.1} & all\_max & 0.23 & 0.17 & 0.23 & 0.087 & 0 & 1.00 & 0.25 \\
 &  & all\_mean & 0.087 & 0.059 & 0.084 & 0.040 & 0 & 0.36 & 0 \\
\midrule
\multirow{10}{*}{ROUGE\_2\_F1} & \multirow{2}{*}{Mistral} & all\_max & 0.062 & 0.017 & 0.048 & 0.070 & 0 & 1.00 & 0 \\
 &  & all\_mean & 0.007 & 0.001 & 0.004 & 0.010 & 0 & 0.29 & 0 \\
\cmidrule{2-10}
 & \multirow{2}{*}{Qwen2.5} & all\_max & 0.064 & 0.010 & 0 & 0.11 & 0 & 1.00 & 0 \\
 &  & all\_mean & 0.006 & 0.001 & 0 & 0.013 & 0 & 0.28 & 0 \\
\cmidrule{2-10}
 & \multirow{2}{*}{Granite} & all\_max & 0.054 & 0.016 & 0.044 & 0.055 & 0 & 1.00 & 0 \\
 &  & all\_mean & 0.007 & 0.001 & 0.004 & 0.009 & 0 & 0.26 & 0 \\
\cmidrule{2-10}
 & \multirow{2}{*}{Falcon3} & all\_max & 0.072 & 0.013 & 0.042 & 0.11 & 0 & 1.00 & 0 \\
 &  & all\_mean & 0.007 & 0.001 & 0.003 & 0.013 & 0 & 0.29 & 0 \\
\cmidrule{2-10}
 & \multirow{2}{*}{Llama-3.1} & all\_max & 0.057 & 0.019 & 0.046 & 0.064 & 0 & 1.00 & 0 \\
 &  & all\_mean & 0.007 & 0.002 & 0.005 & 0.010 & 0 & 0.24 & 0 \\
\midrule
\multirow{10}{*}{ROUGE\_L\_F1} & \multirow{2}{*}{Mistral} & all\_max & 0.18 & 0.12 & 0.17 & 0.078 & 0 & 1.00 & 0.13 \\
 &  & all\_mean & 0.064 & 0.042 & 0.062 & 0.029 & 0 & 0.38 & 0 \\
\cmidrule{2-10}
 & \multirow{2}{*}{Qwen2.5} & all\_max & 0.21 & 0.12 & 0.18 & 0.15 & 0 & 1.00 & 0 \\
 &  & all\_mean & 0.056 & 0.029 & 0.052 & 0.036 & 0 & 0.40 & 0 \\
\cmidrule{2-10}
 & \multirow{2}{*}{Granite} & all\_max & 0.17 & 0.12 & 0.16 & 0.063 & 0 & 1.00 & 0.13 \\
 &  & all\_mean & 0.062 & 0.042 & 0.061 & 0.026 & 0 & 0.35 & 0 \\
\cmidrule{2-10}
 & \multirow{2}{*}{Falcon3} & all\_max & 0.22 & 0.13 & 0.19 & 0.14 & 0 & 1.00 & 0.25 \\
 &  & all\_mean & 0.064 & 0.036 & 0.060 & 0.036 & 0 & 0.39 & 0 \\
\cmidrule{2-10}
 & \multirow{2}{*}{Llama-3.1} & all\_max & 0.17 & 0.12 & 0.16 & 0.076 & 0 & 1.00 & 0.13 \\
 &  & all\_mean & 0.066 & 0.046 & 0.064 & 0.028 & 0 & 0.36 & 0 \\
\bottomrule
\end{tabular}}
\caption{ROUGE metrics: descriptive statistics across the 11{,}528 posts. Mean, median, std, min, max, and mode are for the \emph{real} condition; Mean (rand.) is the random-derangement baseline. Variants: \texttt{all\_max} and \texttt{all\_mean} aggregation over depth-1 replies.}
\label{tab:stats-rouge}
\end{table*}

\begin{table*}[t]
\centering
\footnotesize
\setlength{\tabcolsep}{4pt}
\resizebox{\textwidth}{!}{
\begin{tabular}{lllrrrrrrr}
\toprule
\textbf{Metric} & \textbf{Model} & \textbf{Variant} & \textbf{Mean (real)} & \textbf{Mean (rand.)} & \textbf{Median} & \textbf{Std} & \textbf{Min} & \textbf{Max} & \textbf{Mode} \\
\midrule
\multirow{10}{*}{BERTScore\_F1} & \multirow{2}{*}{Mistral} & all\_max & 0.86 & 0.84 & 0.85 & 0.017 & 0.80 & 1.00 & 0.83 \\
 &  & all\_mean & 0.82 & 0.82 & 0.82 & 0.010 & 0.77 & 0.93 & 0.82 \\
\cmidrule{2-10}
 & \multirow{2}{*}{Qwen2.5} & all\_max & 0.87 & 0.85 & 0.86 & 0.026 & 0.76 & 1.00 & 1.00 \\
 &  & all\_mean & 0.83 & 0.82 & 0.83 & 0.014 & 0.73 & 0.93 & 0.73 \\
\cmidrule{2-10}
 & \multirow{2}{*}{Granite} & all\_max & 0.85 & 0.84 & 0.85 & 0.015 & 0.79 & 0.97 & 0.86 \\
 &  & all\_mean & 0.82 & 0.82 & 0.82 & 0.010 & 0.77 & 0.89 & 0.81 \\
\cmidrule{2-10}
 & \multirow{2}{*}{Falcon3} & all\_max & 0.86 & 0.85 & 0.86 & 0.025 & 0.79 & 1.00 & 1.00 \\
 &  & all\_mean & 0.83 & 0.82 & 0.83 & 0.013 & 0.76 & 0.93 & 0.76 \\
\cmidrule{2-10}
 & \multirow{2}{*}{Llama-3.1} & all\_max & 0.85 & 0.84 & 0.85 & 0.017 & 0.79 & 1.00 & 0.85 \\
 &  & all\_mean & 0.82 & 0.81 & 0.82 & 0.010 & 0.78 & 0.91 & 0.82 \\
\midrule
\multirow{10}{*}{BERTScore\_Precision} & \multirow{2}{*}{Mistral} & all\_max & 0.85 & 0.83 & 0.85 & 0.019 & 0.78 & 1.00 & 0.84 \\
 &  & all\_mean & 0.82 & 0.81 & 0.82 & 0.013 & 0.77 & 0.95 & 0.82 \\
\cmidrule{2-10}
 & \multirow{2}{*}{Qwen2.5} & all\_max & 0.87 & 0.84 & 0.86 & 0.030 & 0.70 & 1.00 & 1.00 \\
 &  & all\_mean & 0.84 & 0.83 & 0.84 & 0.022 & 0.68 & 0.97 & 0.84 \\
\cmidrule{2-10}
 & \multirow{2}{*}{Granite} & all\_max & 0.85 & 0.83 & 0.85 & 0.017 & 0.77 & 0.98 & 0.81 \\
 &  & all\_mean & 0.82 & 0.81 & 0.82 & 0.013 & 0.75 & 0.95 & 0.81 \\
\cmidrule{2-10}
 & \multirow{2}{*}{Falcon3} & all\_max & 0.86 & 0.84 & 0.86 & 0.028 & 0.75 & 1.00 & 1.00 \\
 &  & all\_mean & 0.83 & 0.82 & 0.83 & 0.020 & 0.73 & 0.96 & 0.73 \\
\cmidrule{2-10}
 & \multirow{2}{*}{Llama-3.1} & all\_max & 0.85 & 0.83 & 0.85 & 0.019 & 0.77 & 1.00 & 1.00 \\
 &  & all\_mean & 0.82 & 0.81 & 0.82 & 0.015 & 0.74 & 0.96 & 0.74 \\
\midrule
\multirow{10}{*}{BERTScore\_Recall} & \multirow{2}{*}{Mistral} & all\_max & 0.88 & 0.87 & 0.88 & 0.028 & 0.81 & 1.00 & 0.84 \\
 &  & all\_mean & 0.83 & 0.82 & 0.83 & 0.011 & 0.77 & 0.91 & 0.83 \\
\cmidrule{2-10}
 & \multirow{2}{*}{Qwen2.5} & all\_max & 0.89 & 0.87 & 0.88 & 0.034 & 0.81 & 1.00 & 1.00 \\
 &  & all\_mean & 0.83 & 0.82 & 0.83 & 0.012 & 0.77 & 0.92 & 0.81 \\
\cmidrule{2-10}
 & \multirow{2}{*}{Granite} & all\_max & 0.88 & 0.87 & 0.87 & 0.028 & 0.80 & 1.00 & 0.84 \\
 &  & all\_mean & 0.83 & 0.82 & 0.83 & 0.011 & 0.76 & 0.89 & 0.81 \\
\cmidrule{2-10}
 & \multirow{2}{*}{Falcon3} & all\_max & 0.89 & 0.87 & 0.88 & 0.033 & 0.80 & 1.00 & 1.00 \\
 &  & all\_mean & 0.83 & 0.82 & 0.83 & 0.012 & 0.77 & 0.91 & 0.84 \\
\cmidrule{2-10}
 & \multirow{2}{*}{Llama-3.1} & all\_max & 0.88 & 0.87 & 0.88 & 0.030 & 0.80 & 1.00 & 1.00 \\
 &  & all\_mean & 0.83 & 0.82 & 0.83 & 0.012 & 0.76 & 0.89 & 0.83 \\
\bottomrule
\end{tabular}}
\caption{BERTScore (raw) metrics: descriptive statistics across the 11{,}528 posts. Mean, median, std, min, max, and mode are for the \emph{real} condition; Mean (rand.) is the random-derangement baseline. Variants: \texttt{all\_max} and \texttt{all\_mean} aggregation over depth-1 replies.}
\label{tab:stats-bertscore_raw}
\end{table*}

\begin{table*}[t]
\centering
\footnotesize
\setlength{\tabcolsep}{4pt}
\resizebox{\textwidth}{!}{
\begin{tabular}{lllrrrrrrr}
\toprule
\textbf{Metric} & \textbf{Model} & \textbf{Variant} & \textbf{Mean (real)} & \textbf{Mean (rand.)} & \textbf{Median} & \textbf{Std} & \textbf{Min} & \textbf{Max} & \textbf{Mode} \\
\midrule
\multirow{10}{*}{BERTScore\_F1\_Rescaled} & \multirow{2}{*}{Mistral} & all\_max & 0.21 & 0.12 & 0.20 & 0.092 & -0.083 & 1.00 & 0.094 \\
 &  & all\_mean & 0.041 & -0.007 & 0.042 & 0.056 & -0.24 & 0.60 & -0.24 \\
\cmidrule{2-10}
 & \multirow{2}{*}{Qwen2.5} & all\_max & 0.27 & 0.18 & 0.24 & 0.14 & -0.32 & 1.00 & 1.00 \\
 &  & all\_mean & 0.078 & 0.026 & 0.077 & 0.076 & -0.48 & 0.61 & -0.48 \\
\cmidrule{2-10}
 & \multirow{2}{*}{Granite} & all\_max & 0.20 & 0.12 & 0.19 & 0.084 & -0.17 & 0.83 & 0.21 \\
 &  & all\_mean & 0.034 & -0.012 & 0.037 & 0.055 & -0.26 & 0.41 & -0.26 \\
\cmidrule{2-10}
 & \multirow{2}{*}{Falcon3} & all\_max & 0.25 & 0.16 & 0.23 & 0.14 & -0.15 & 1.00 & 1.00 \\
 &  & all\_mean & 0.067 & 0.015 & 0.064 & 0.071 & -0.32 & 0.62 & -0.32 \\
\cmidrule{2-10}
 & \multirow{2}{*}{Llama-3.1} & all\_max & 0.20 & 0.12 & 0.19 & 0.094 & -0.13 & 1.00 & 0.20 \\
 &  & all\_mean & 0.036 & -0.015 & 0.037 & 0.057 & -0.23 & 0.52 & -0.23 \\
\midrule
\multirow{10}{*}{BERTScore\_Precision\_Rescaled} & \multirow{2}{*}{Mistral} & all\_max & 0.19 & 0.080 & 0.18 & 0.10 & -0.21 & 1.00 & 0.10 \\
 &  & all\_mean & 0.027 & -0.015 & 0.027 & 0.072 & -0.28 & 0.71 & -0.28 \\
\cmidrule{2-10}
 & \multirow{2}{*}{Qwen2.5} & all\_max & 0.27 & 0.14 & 0.25 & 0.16 & -0.64 & 1.00 & 1.00 \\
 &  & all\_mean & 0.11 & 0.056 & 0.10 & 0.12 & -0.78 & 0.81 & -0.78 \\
\cmidrule{2-10}
 & \multirow{2}{*}{Granite} & all\_max & 0.17 & 0.069 & 0.17 & 0.094 & -0.28 & 0.90 & -0.051 \\
 &  & all\_mean & 0.014 & -0.024 & 0.017 & 0.070 & -0.35 & 0.75 & -0.35 \\
\cmidrule{2-10}
 & \multirow{2}{*}{Falcon3} & all\_max & 0.25 & 0.12 & 0.23 & 0.15 & -0.37 & 1.00 & 1.00 \\
 &  & all\_mean & 0.081 & 0.029 & 0.075 & 0.11 & -0.50 & 0.80 & -0.50 \\
\cmidrule{2-10}
 & \multirow{2}{*}{Llama-3.1} & all\_max & 0.17 & 0.064 & 0.16 & 0.10 & -0.27 & 1.00 & 1.00 \\
 &  & all\_mean & 0.007 & -0.034 & 0.005 & 0.080 & -0.42 & 0.78 & -0.42 \\
\midrule
\multirow{10}{*}{BERTScore\_Recall\_Rescaled} & \multirow{2}{*}{Mistral} & all\_max & 0.35 & 0.28 & 0.32 & 0.15 & -0.041 & 1.00 & 0.14 \\
 &  & all\_mean & 0.056 & 0.002 & 0.056 & 0.061 & -0.29 & 0.50 & -0.29 \\
\cmidrule{2-10}
 & \multirow{2}{*}{Qwen2.5} & all\_max & 0.38 & 0.31 & 0.34 & 0.19 & -0.020 & 1.00 & 1.00 \\
 &  & all\_mean & 0.050 & -0.002 & 0.048 & 0.068 & -0.25 & 0.58 & -0.25 \\
\cmidrule{2-10}
 & \multirow{2}{*}{Granite} & all\_max & 0.35 & 0.27 & 0.31 & 0.15 & -0.076 & 1.00 & 0.12 \\
 &  & all\_mean & 0.055 & 0.001 & 0.055 & 0.061 & -0.30 & 0.41 & -0.30 \\
\cmidrule{2-10}
 & \multirow{2}{*}{Falcon3} & all\_max & 0.38 & 0.31 & 0.34 & 0.18 & -0.077 & 1.00 & 1.00 \\
 &  & all\_mean & 0.055 & 0.003 & 0.052 & 0.067 & -0.28 & 0.51 & -0.28 \\
\cmidrule{2-10}
 & \multirow{2}{*}{Llama-3.1} & all\_max & 0.37 & 0.28 & 0.33 & 0.16 & -0.072 & 1.00 & 1.00 \\
 &  & all\_mean & 0.066 & 0.005 & 0.066 & 0.063 & -0.29 & 0.40 & -0.29 \\
\bottomrule
\end{tabular}}
\caption{BERTScore (baseline-rescaled) metrics: descriptive statistics across the 11{,}528 posts. Mean, median, std, min, max, and mode are for the \emph{real} condition; Mean (rand.) is the random-derangement baseline. Variants: \texttt{all\_max} and \texttt{all\_mean} aggregation over depth-1 replies.}
\label{tab:stats-bertscore_rescaled}
\end{table*}

\begin{table*}[t]
\centering
\footnotesize
\setlength{\tabcolsep}{4pt}
\resizebox{\textwidth}{!}{
\begin{tabular}{lllrrrrrrr}
\toprule
\textbf{Metric} & \textbf{Model} & \textbf{Variant} & \textbf{Mean (real)} & \textbf{Mean (rand.)} & \textbf{Median} & \textbf{Std} & \textbf{Min} & \textbf{Max} & \textbf{Mode} \\
\midrule
\multirow{10}{*}{Cosine\_Similarity\_minilm} & \multirow{2}{*}{Mistral} & all\_max & 0.54 & 0.19 & 0.55 & 0.15 & 0.057 & 1.00 & 0.24 \\
 &  & all\_mean & 0.25 & 0.048 & 0.25 & 0.093 & -0.026 & 0.70 & -0.026 \\
\cmidrule{2-10}
 & \multirow{2}{*}{Qwen2.5} & all\_max & 0.51 & 0.21 & 0.51 & 0.16 & 0.082 & 1.00 & 1.00 \\
 &  & all\_mean & 0.25 & 0.065 & 0.24 & 0.094 & -0.036 & 0.71 & -0.036 \\
\cmidrule{2-10}
 & \multirow{2}{*}{Granite} & all\_max & 0.53 & 0.19 & 0.54 & 0.15 & 0.081 & 0.95 & 0.46 \\
 &  & all\_mean & 0.25 & 0.046 & 0.24 & 0.094 & -0.011 & 0.68 & -0.011 \\
\cmidrule{2-10}
 & \multirow{2}{*}{Falcon3} & all\_max & 0.53 & 0.21 & 0.52 & 0.16 & 0.082 & 1.00 & 1.00 \\
 &  & all\_mean & 0.25 & 0.061 & 0.24 & 0.094 & -0.020 & 0.72 & -0.020 \\
\cmidrule{2-10}
 & \multirow{2}{*}{Llama-3.1} & all\_max & 0.52 & 0.20 & 0.53 & 0.15 & 0.035 & 1.00 & 1.00 \\
 &  & all\_mean & 0.25 & 0.052 & 0.25 & 0.095 & -0.051 & 0.72 & -0.051 \\
\midrule
\multirow{10}{*}{Cosine\_Similarity\_mpnet} & \multirow{2}{*}{Mistral} & all\_max & 0.59 & 0.17 & 0.60 & 0.15 & 0.058 & 1.00 & 0.18 \\
 &  & all\_mean & 0.29 & 0.040 & 0.28 & 0.100 & -0.030 & 0.71 & -0.030 \\
\cmidrule{2-10}
 & \multirow{2}{*}{Qwen2.5} & all\_max & 0.56 & 0.20 & 0.56 & 0.16 & 0.071 & 1.00 & 1.00 \\
 &  & all\_mean & 0.29 & 0.056 & 0.28 & 0.100 & -0.009 & 0.76 & -0.009 \\
\cmidrule{2-10}
 & \multirow{2}{*}{Granite} & all\_max & 0.58 & 0.17 & 0.59 & 0.15 & 0.085 & 0.97 & 0.28 \\
 &  & all\_mean & 0.29 & 0.040 & 0.28 & 0.100 & -0.013 & 0.70 & -0.013 \\
\cmidrule{2-10}
 & \multirow{2}{*}{Falcon3} & all\_max & 0.57 & 0.19 & 0.58 & 0.16 & 0.034 & 1.00 & 1.00 \\
 &  & all\_mean & 0.29 & 0.053 & 0.29 & 0.10 & -0.042 & 0.77 & -0.042 \\
\cmidrule{2-10}
 & \multirow{2}{*}{Llama-3.1} & all\_max & 0.58 & 0.18 & 0.59 & 0.15 & 0.040 & 1.00 & 1.00 \\
 &  & all\_mean & 0.29 & 0.046 & 0.29 & 0.10 & -0.044 & 0.77 & -0.044 \\
\bottomrule
\end{tabular}}
\caption{Cosine Similarity metrics: descriptive statistics across the 11{,}528 posts. Mean, median, std, min, max, and mode are for the \emph{real} condition; Mean (rand.) is the random-derangement baseline. Variants: \texttt{all\_max} and \texttt{all\_mean} aggregation over depth-1 replies.}
\label{tab:stats-cosine}
\end{table*}

\begin{table*}[t]
\centering
\footnotesize
\setlength{\tabcolsep}{4pt}
\resizebox{\textwidth}{!}{
\begin{tabular}{lllrrrrrrr}
\toprule
\textbf{Metric} & \textbf{Model} & \textbf{Variant} & \textbf{Mean (real)} & \textbf{Mean (rand.)} & \textbf{Median} & \textbf{Std} & \textbf{Min} & \textbf{Max} & \textbf{Mode} \\
\midrule
\multirow{10}{*}{MoverScore} & \multirow{2}{*}{Mistral} & all\_max & 0.12 & 0.015 & 0.11 & 0.092 & -0.15 & 1.00 & 1.00 \\
 &  & all\_mean & -0.016 & -0.068 & -0.016 & 0.044 & -0.19 & 0.24 & -0.19 \\
\cmidrule{2-10}
 & \multirow{2}{*}{Qwen2.5} & all\_max & 0.13 & 0.016 & 0.11 & 0.13 & -0.14 & 1.00 & 1.00 \\
 &  & all\_mean & -0.008 & -0.065 & -0.009 & 0.045 & -0.19 & 1.00 & -0.19 \\
\cmidrule{2-10}
 & \multirow{2}{*}{Granite} & all\_max & 0.11 & 0.012 & 0.10 & 0.089 & -0.098 & 1.00 & 1.00 \\
 &  & all\_mean & -0.021 & -0.071 & -0.021 & 0.044 & -0.18 & 0.17 & -0.18 \\
\cmidrule{2-10}
 & \multirow{2}{*}{Falcon3} & all\_max & 0.14 & 0.016 & 0.11 & 0.13 & -0.100 & 1.00 & 1.00 \\
 &  & all\_mean & -0.010 & -0.066 & -0.012 & 0.045 & -0.18 & 0.27 & -0.18 \\
\cmidrule{2-10}
 & \multirow{2}{*}{Llama-3.1} & all\_max & 0.12 & 0.018 & 0.11 & 0.094 & -0.13 & 1.00 & 1.00 \\
 &  & all\_mean & -0.015 & -0.067 & -0.015 & 0.045 & -0.19 & 0.22 & -0.19 \\
\bottomrule
\end{tabular}}
\caption{MoverScore metrics: descriptive statistics across the 11{,}528 posts. Mean, median, std, min, max, and mode are for the \emph{real} condition; Mean (rand.) is the random-derangement baseline. Variants: \texttt{all\_max} and \texttt{all\_mean} aggregation over depth-1 replies.}
\label{tab:stats-moverscore}
\end{table*}

\begin{table*}[t]
\centering
\footnotesize
\setlength{\tabcolsep}{4pt}
\resizebox{\textwidth}{!}{
\begin{tabular}{lllrrrrrrr}
\toprule
\textbf{Metric} & \textbf{Model} & \textbf{Variant} & \textbf{Mean (real)} & \textbf{Mean (rand.)} & \textbf{Median} & \textbf{Std} & \textbf{Min} & \textbf{Max} & \textbf{Mode} \\
\midrule
\multirow{10}{*}{NLI\_Contradiction\_Pct} & \multirow{2}{*}{Mistral} & all\_max & 0.56 & 0.43 & 0.50 & 0.42 & 0 & 1.00 & 1.00 \\
 &  & all\_mean & 0.22 & 0.16 & 0.11 & 0.26 & 0 & 1.00 & 0 \\
\cmidrule{2-10}
 & \multirow{2}{*}{Qwen2.5} & all\_max & 0.77 & 0.68 & 1.00 & 0.40 & 0 & 1.00 & 1.00 \\
 &  & all\_mean & 0.36 & 0.32 & 0.28 & 0.32 & 0 & 1.00 & 0 \\
\cmidrule{2-10}
 & \multirow{2}{*}{Granite} & all\_max & 0.59 & 0.51 & 0.50 & 0.35 & 0 & 1.00 & 1.00 \\
 &  & all\_mean & 0.24 & 0.20 & 0.18 & 0.23 & 0 & 1.00 & 0 \\
\cmidrule{2-10}
 & \multirow{2}{*}{Falcon3} & all\_max & 0.76 & 0.63 & 1.00 & 0.42 & 0 & 1.00 & 1.00 \\
 &  & all\_mean & 0.34 & 0.28 & 0.25 & 0.33 & 0 & 1.00 & 0 \\
\cmidrule{2-10}
 & \multirow{2}{*}{Llama-3.1} & all\_max & 0.66 & 0.57 & 0.67 & 0.33 & 0 & 1.00 & 1.00 \\
 &  & all\_mean & 0.28 & 0.25 & 0.24 & 0.23 & 0 & 1.00 & 0 \\
\midrule
\multirow{10}{*}{NLI\_Entailment\_Pct} & \multirow{2}{*}{Mistral} & all\_max & 0.19 & 0.034 & 0 & 0.33 & 0 & 1.00 & 0 \\
 &  & all\_mean & 0.021 & 0.003 & 0 & 0.056 & 0 & 0.97 & 0 \\
\cmidrule{2-10}
 & \multirow{2}{*}{Qwen2.5} & all\_max & 0.30 & 0.067 & 0 & 0.44 & 0 & 1.00 & 0 \\
 &  & all\_mean & 0.036 & 0.006 & 0 & 0.080 & 0 & 1.00 & 0 \\
\cmidrule{2-10}
 & \multirow{2}{*}{Granite} & all\_max & 0.19 & 0.050 & 0 & 0.28 & 0 & 1.00 & 0 \\
 &  & all\_mean & 0.023 & 0.005 & 0 & 0.047 & 0 & 0.86 & 0 \\
\cmidrule{2-10}
 & \multirow{2}{*}{Falcon3} & all\_max & 0.27 & 0.052 & 0 & 0.44 & 0 & 1.00 & 0 \\
 &  & all\_mean & 0.032 & 0.006 & 0 & 0.078 & 0 & 1.00 & 0 \\
\cmidrule{2-10}
 & \multirow{2}{*}{Llama-3.1} & all\_max & 0.16 & 0.034 & 0 & 0.27 & 0 & 1.00 & 0 \\
 &  & all\_mean & 0.019 & 0.003 & 0 & 0.047 & 0 & 0.81 & 0 \\
\midrule
\multirow{10}{*}{NLI\_Neutral\_Pct} & \multirow{2}{*}{Mistral} & all\_max & 0.99 & 0.98 & 1.00 & 0.10 & 0 & 1.00 & 1.00 \\
 &  & all\_mean & 0.76 & 0.84 & 0.85 & 0.26 & 0 & 1.00 & 1.00 \\
\cmidrule{2-10}
 & \multirow{2}{*}{Qwen2.5} & all\_max & 0.96 & 0.95 & 1.00 & 0.20 & 0 & 1.00 & 1.00 \\
 &  & all\_mean & 0.60 & 0.67 & 0.67 & 0.32 & 0 & 1.00 & 1.00 \\
\cmidrule{2-10}
 & \multirow{2}{*}{Granite} & all\_max & 0.97 & 0.97 & 1.00 & 0.11 & 0 & 1.00 & 1.00 \\
 &  & all\_mean & 0.74 & 0.80 & 0.79 & 0.23 & 0 & 1.00 & 1.00 \\
\cmidrule{2-10}
 & \multirow{2}{*}{Falcon3} & all\_max & 0.97 & 0.97 & 1.00 & 0.18 & 0 & 1.00 & 1.00 \\
 &  & all\_mean & 0.63 & 0.72 & 0.70 & 0.32 & 0 & 1.00 & 1.00 \\
\cmidrule{2-10}
 & \multirow{2}{*}{Llama-3.1} & all\_max & 0.97 & 0.96 & 1.00 & 0.13 & 0 & 1.00 & 1.00 \\
 &  & all\_mean & 0.70 & 0.75 & 0.73 & 0.23 & 0 & 1.00 & 1.00 \\
\bottomrule
\end{tabular}}
\caption{NLI metrics: descriptive statistics across the 11{,}528 posts. Mean, median, std, min, max, and mode are for the \emph{real} condition; Mean (rand.) is the random-derangement baseline. Variants: \texttt{all\_max} and \texttt{all\_mean} aggregation over depth-1 replies.}
\label{tab:stats-nli}
\end{table*}

\begin{table*}[t]
\centering
\footnotesize
\setlength{\tabcolsep}{4pt}
\resizebox{\textwidth}{!}{
\begin{tabular}{lllrrrrrrr}
\toprule
\textbf{Metric} & \textbf{Model} & \textbf{Variant} & \textbf{Mean (real)} & \textbf{Mean (rand.)} & \textbf{Median} & \textbf{Std} & \textbf{Min} & \textbf{Max} & \textbf{Mode} \\
\midrule
\multirow{10}{*}{Keyword\_Overlap} & \multirow{2}{*}{Mistral} & all\_max & 0.067 & 0.027 & 0.059 & 0.050 & 0 & 1.00 & 0 \\
 &  & all\_mean & 0.012 & 0.003 & 0.009 & 0.012 & 0 & 0.18 & 0 \\
\cmidrule{2-10}
 & \multirow{2}{*}{Qwen2.5} & all\_max & 0.080 & 0.022 & 0.062 & 0.10 & 0 & 1.00 & 0 \\
 &  & all\_mean & 0.011 & 0.002 & 0.007 & 0.015 & 0 & 0.22 & 0 \\
\cmidrule{2-10}
 & \multirow{2}{*}{Granite} & all\_max & 0.059 & 0.025 & 0.053 & 0.039 & 0 & 0.64 & 0 \\
 &  & all\_mean & 0.011 & 0.003 & 0.009 & 0.010 & 0 & 0.11 & 0 \\
\cmidrule{2-10}
 & \multirow{2}{*}{Falcon3} & all\_max & 0.083 & 0.024 & 0.062 & 0.10 & 0 & 1.00 & 0 \\
 &  & all\_mean & 0.012 & 0.002 & 0.008 & 0.016 & 0 & 0.33 & 0 \\
\cmidrule{2-10}
 & \multirow{2}{*}{Llama-3.1} & all\_max & 0.063 & 0.029 & 0.056 & 0.050 & 0 & 1.00 & 0 \\
 &  & all\_mean & 0.013 & 0.004 & 0.010 & 0.011 & 0 & 0.18 & 0 \\
\midrule
\multirow{10}{*}{Length\_Ratio} & \multirow{2}{*}{Mistral} & all\_max & 21.9 & 22.3 & 21.0 & 12.3 & 0.83 & 122.0 & 25.0 \\
 &  & all\_mean & 6.06 & 6.42 & 5.05 & 4.06 & 0.26 & 45.7 & 0.26 \\
\cmidrule{2-10}
 & \multirow{2}{*}{Qwen2.5} & all\_max & 9.33 & 9.66 & 8.50 & 6.01 & 0.20 & 50.0 & 10.0 \\
 &  & all\_mean & 2.50 & 2.78 & 2.05 & 1.75 & 0.079 & 17.5 & 0.83 \\
\cmidrule{2-10}
 & \multirow{2}{*}{Granite} & all\_max & 26.0 & 26.4 & 26.0 & 13.9 & 0.75 & 198.0 & 28.0 \\
 &  & all\_mean & 7.25 & 7.57 & 6.19 & 4.72 & 0.43 & 63.9 & 10.3 \\
\cmidrule{2-10}
 & \multirow{2}{*}{Falcon3} & all\_max & 12.0 & 12.3 & 11.0 & 8.06 & 0.24 & 66.0 & 8.00 \\
 &  & all\_mean & 3.25 & 3.51 & 2.58 & 2.52 & 0.13 & 21.8 & 1.07 \\
\cmidrule{2-10}
 & \multirow{2}{*}{Llama-3.1} & all\_max & 26.4 & 26.6 & 29.0 & 12.5 & 0.33 & 128.0 & 36.0 \\
 &  & all\_mean & 7.37 & 7.68 & 6.48 & 4.47 & 0.17 & 66.4 & 5.79 \\
\bottomrule
\end{tabular}}
\caption{Length and Lexical metrics: descriptive statistics across the 11{,}528 posts. Mean, median, std, min, max, and mode are for the \emph{real} condition; Mean (rand.) is the random-derangement baseline. Variants: \texttt{all\_max} and \texttt{all\_mean} aggregation over depth-1 replies.}
\label{tab:stats-length_quality}
\end{table*}

\end{document}